\documentclass[10pt,twocolumn,twoside]{IEEEtran}

\usepackage[pdftex]{graphicx}
\graphicspath{{figure/}}
\usepackage[dvipsnames]{xcolor}
\usepackage{tikz}

\usepackage[T1]{fontenc}
\usepackage{amsmath}
\usepackage{mathtools}
\usepackage{amsfonts}
\usepackage{bm}
\usepackage{graphicx}
\usepackage{caption}
\usepackage{subcaption}
\usepackage{xcolor,colortbl}
\usepackage{siunitx}
\usepackage{url}

\usepackage{algorithm}
\usepackage{algpseudocode}
\usepackage{multirow}
\usepackage{verbatim}
\usepackage{amssymb}
\usepackage{soul}

\usepackage{array}

\newcolumntype{C}[1]{>{\centering\arraybackslash}m{#1}}
\newcolumntype{L}[1]{>{\arraybackslash}m{#1}}

\newcommand{\ournet}{\textsf{4-Twins Net}}
\newcommand{\oursiam}{\textsf{Siamese Net}}
\newcommand{\bP}{\mathcal{P}}
\newcommand{\bQ}{\mathcal{Q}}

\DeclarePairedDelimiter\floor{\lfloor}{\rfloor}

\colorlet{mygreen}{green!60!gray}

\newcommand{\CORR}[1]{\textcolor{black}{{#1}}}
\newcommand{\BTT}[1]{\textcolor{black}{{#1}}}

\begin{document}

\title{Copy Move Source-Target Disambiguation through Multi-Branch CNNs}

\author{*Mauro~Barni, \IEEEmembership{Fellow, IEEE}, Quoc-Tin~Phan, Benedetta~Tondi, \IEEEmembership{Member, IEEE}%
\thanks{Mauro~Barni and Benedetta~Tondi  are with the Department of  Information Engineering and Mathematics, University of Siena, 53100 Siena, ITALY; Quoc-Tin~Phan is with the Department of  Information Engineering and  Computer Science,  University of Trento, 38123 Trento, ITALY.\newline
* The list of authors is provided in alphabetic order.}
\vspace{-0.5cm}
}

\maketitle

\IEEEtitleabstractindextext{

\begin{abstract}
We propose a method to identify the source and target regions of a copy-move forgery  so allow a correct localisation of  the tampered area.
First, we cast the problem into a hypothesis testing framework whose goal is to decide which region between the two nearly-duplicate regions detected by a generic copy-move detector is the original one. Then we design a multi-branch CNN architecture that solves the hypothesis testing problem by learning a set of features capable to reveal the presence of interpolation artefacts and boundary inconsistencies in the copy-moved area.
The proposed architecture, trained on a synthetic dataset explicitly built for this purpose, achieves good results on copy-move forgeries from both synthetic and realistic datasets. Based on our tests, the proposed disambiguation method  can reliably reveal the target region even in realistic cases where an approximate version of the copy-move localization mask is provided by a state-of-the-art copy-move detection algorithm.
\end{abstract}
\begin{IEEEkeywords}
Copy-move detection and localization, image forensics, tampering detection and localization, deep learning for forensics, Siamese networks.
\end{IEEEkeywords}
}

\IEEEdisplaynontitleabstractindextext
\IEEEpeerreviewmaketitle

%
%

\section{Introduction}

Thanks to the wide availability of easy-to-use image editing tools,  altering the visual content of digital images is becoming simpler and simpler.
Copy-Move (CM) forgery, where an image region is copied into another part of the same image, is one of the most common and easy-to-implement image tampering techniques. To detect this kind of forgery, several CM detection and localization algorithms have been proposed, attempting to determine whether a given image contains cloned regions, or so called {\em nearly duplicate regions} (in which case, the image is  labeled as a suspect or forged image).
The great majority of the algorithms proposed so far rely on local hand-crafted features \cite{RiessTIFS,tan2019survey},
and are grouped into two main categories: block-based (also called patch-based) methods, e.g. \cite{fridrich2003detection,Cozzolino2015}, and keypoints-based methods,
\cite{Huang2008,Amerini2011,SilvaSURF2015}.
%
Both approaches have their
strengths and weaknesses
and a solution capable to outperform all the others in every working conditions is not available yet.
%
Motivated by the recent
trend towards the adoption of Deep Learning (DL) methods for image forensic tasks,  DL-based approaches have also been proposed for CM detection. Such methods are capable to automatically learn and extract descriptors from the image, e.g. in \cite{rao2016deep,liu2018copy}, by means of Deep Neural Network (DNNs), that hence work as feature extractors.
End-to-end DNN-based solutions for copy-move tampering localization have also been proposed, as in \cite{wu2018image, Wu2018}, where a convolutional and de-convolutional module work together to directly produce a copy move forgery mask from the to-be-analyzed input image.

The great majority of the algorithms proposed so far
can only detect the copy-move forgery and localize the nearly duplicate areas, providing a binary mask that highlights both the source region and its displaced version, without identifying which of the two regions corresponds to the source area and which to the target one. However, in hindsight,  only the target region of a copy-move forgery corresponds to a manipulated area; therefore, distinguishing  between source and target regions is of primary importance to correctly localize the tampered area  and possibly trace back to the goal of the forgery.
%
To the best of our knowledge, the only paper addressing the problem of source-target disambiguation in general copy-move forgeries is \cite{Wu2018}. In that work, an end-to-end system for CM localization and disambiguation, called BusterNet, is proposed, based on a DNN architecture with two-branches. The first branch is designed to extract a pool of features revealing  general traces of manipulations. These features are then combined with those extracted from the other branch, in charge of copy-move detection.
With regard to source-target disambiguation, however,
the performance achieved by the method on realistic publicly available CM datasets are rather limited.
As stated by the authors themselves, this may be due to
the limited performance of the manipulation detection branch, which  tends to overfit to the synthetic dataset used for training.

%

In this paper, we propose a new DNN-based method to address the problem of source-target disambiguation in images subject to CM manipulation.
Given the binary localization mask produced by a generic copy-move detector, our method permits to derive the actual tampering mask, by identifying the target and source region of the copy-move.
The main idea behind the proposed method is to exploit the {\em non-invertibility} of the copy-move transformation, due to the presence of interpolation artefacts and local post-processing traces in the displaced region.
Specifically, we propose a multi-branch CNN architecture, called \textsf{DisTool}, consisting of two main parallel branches, looking for two different kinds of CM-traces. The first branch, named {\em 4-Twins Net}, consists of two parallel Siamese networks, trained in such a way to exploit the non-invertibility of the copy-move process caused by the {\em interpolation} artefacts often associated to the copy-move operation. The second branch 
is a Siamese network \cite{Bromley1993} designed to identify artefacts and inconsistencies present at the boundary of the copy-moved region.
%
%
%
The soft outputs of the two branches are, finally, fused through a simple fusion module.
A remarkable strength of the proposed method is that it works independently of the CM detection algorithm, and hence it can be used on top of any such method.
\CORR{In this way, the system designer has the freedom to choose the CM algorithm that best suits the application at hand (for instance, it is known that SIFT-based approaches work very well when the size of the copy-moved area  is large, while performance drop with small regions).
The difficulty of training an end-to-end architecture for both localization and disambiguation providing  good performance on both tasks also motivates the use of an independent tool for the disambiguation.}
Our experiments show that the proposed method has a very good disambiguation capabilities, greatly outperforming those of \cite{Wu2018}, and that it generalizes well to both synthetic and realistic copy-move forgeries from several different datasets.
Robustness to post-processing is also good.

The paper is organized as follows. In Section \ref{sec.formulation}, we formalize the CM source-target disambiguation problem addressed in the paper, and present the rationale behind the proposed method.
%
The details of the multi-branch CNNs composing the system are given in Section \ref{sec.multibrachCNN}. In Section \ref{sec.methods}, we describe the methodology we followed to run the experiments whereby we validated the effectiveness of the proposed method. The results of the experiments are reported and discussed in Section \ref{sec.results}. The paper ends in Section \ref{sec.conc}, with some concluding remarks.

\section{Problem formulation and overall description of the proposed method}
\label{sec.formulation}

In this section, we provide a rigorous formulation of the source-target disambiguation problem and present the overall architecture of the proposed system.
Before doing that, we introduce some basic concepts and notation, and detail the main steps involved in the creation of a copy-move forgery.

Among  the various instances of copy-move forgeries that can be encountered in practice,
in this work, we focus on the common, and simplest, case of a single source region copy-moved into a single target location (referred to as (1-1) CM).
The case of $n$ sources singularly copied into $n$ target locations, namely the ($n$-$n$) case, can be interpreted as multiple instances of the (1-1) case and can be treated as such.
\BTT{When the target region is partially overlapped to the source, only the non-overlapping parts of the copied and pasted regions are regarded to as a copy-move forgery. In this case, the proposed system can be straightforwardly applied if the CM algorithm returns the two nearly-duplicate regions (as it is the case for instance with keypoint-based methods). When a unique region is returned by the CM algorithm, segmentation methods must be applied to split it into to two nearly-duplicate regions.}


\begin{figure}[h!]
	\centering
	\begin{subfigure}{0.24\linewidth}
		\centering
		\includegraphics[width=\linewidth]{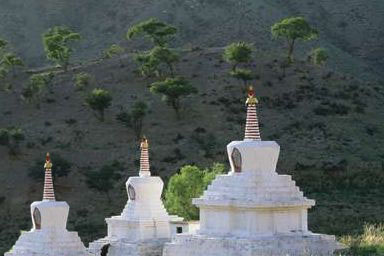}
		\caption{}
	\end{subfigure}
	\begin{subfigure}{0.24\linewidth}
		\centering
		\includegraphics[width=\linewidth]{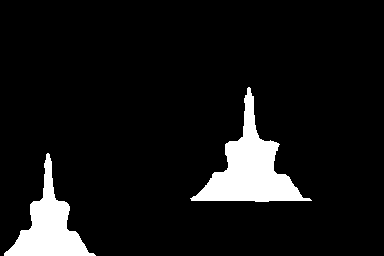}
		\caption{}
	\end{subfigure}
\begin{subfigure}{0.24\linewidth}
		\centering
		\includegraphics[width=\linewidth]{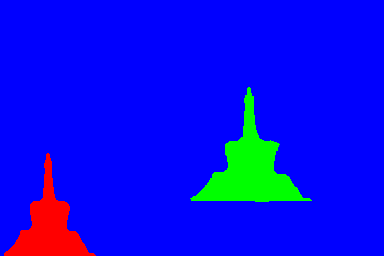}
		\caption{}
	\end{subfigure}
\begin{subfigure}{0.24\linewidth}
		\centering
		\includegraphics[width=\linewidth]{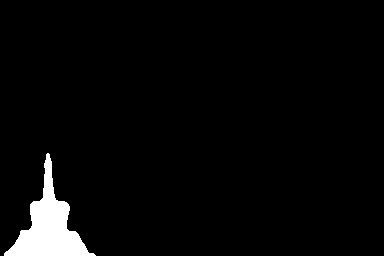}
		\caption{}
	\end{subfigure}
	\caption{A  CM forged image (a); corresponding binary localization mask  (b);  tampering map with highlighted the source (green)
 and target (red) regions (c);  final (desired) tampering mask (d).}
	\label{fig:cm_example}
\end{figure}

\subsection{Preliminaries}
\label{sec.preliminaries}

Let $I$ be the original image of  size $l\times m$, and  $I_f$ the copy-move forgery, of the same size\footnote{For simplicity,  we consider the case of gray-scale images. Similar arguments apply to the case of color images.}.
%
We denote with $S$ and $T$ the subparts of $I_f$ corresponding to the source and the target region, respectively.
%
%

During a copy-move forgery, the region $S$ is copied, possibly after a geometric transformation, and pasted into $T$.
In its most basic form, the copy-move operation can be modelled as a geometric affine transformation between $S$ and $T$. Let $\mathcal{H}_\theta$ denote the transformation that maps a generic point $(u,v)$ in the source region $S$ into another point $(u',v')$ in $T$, parameterized by a vector $\theta$.
%
%
%
%
%
%
Such a transformation can be represented by a matrix $H_{\theta}$,
that represents a rotation, resizing or scaling, sheering, translation, or, more in general, a composition of them.
%
Then, ideally, for every $(u',v')$, we would have
$I_f(u',v') = I(u,v)$, 
where the relation between  $(u,v)$  and $(u',v')$ is established by the matrix $H_{\theta}$.
In general, after the transformation, the mapped point is not a valid point in the 2D regular pixel grid,
%
%
and the pixel values at the regular grid points  are obtained by interpolating the neighboring pixels of the source region by means of a kernel function  $k(\cdot,\cdot)$\CORR{ \cite{gonzales2002digital}}.
%
%
%
%
%
We let $\Psi_{\theta,k}$ denote the transformation that maps the pixels in the source regions $S$ to those in the target region $T$, taking into account both the geometric transformation $H_\theta$ and the interpolation process with kernel $k$. Then, $T = \Psi_{\theta,k}(S)$\footnote{Strictly speaking, some pixels on the boundary of $T$ are obtained by interpolating also pixels that do not belong to $S$, i.e. pixels in $\bar{S}$.}.
The interpolation process introduces correlations among neighboring pixels in $T$.
After interpolation, $S$ and $T$ are {\em nearly duplicate} regions (the regions are not identical because of the interpolation).
%
%
%
In most cases, the interpolation process makes the copy-move operation \textit{non-invertible}.
\CORR{As a matter of fact, interpolation traces have been widely studied by several multimedia forensics works for resampling detection and more in general forgery localization, e.g.  \cite{kirchner2008fast,blindAutentication, Popescu, FernandoCM}.}

In realistic copy-move forgeries, various post-processing operations might also be applied locally to the target region in order to hide the traces of copy-pasting. For example, the pasted region and the background are often blended to visually hide the transition from the copied part and the surrounding area. Post-processing might also be applied globally, in which case it affects both the source and the target regions.

An example of CM forged image is provided in Fig. \ref{fig:cm_example}(a) along with the corresponding localization mask (b). The disambiguation map is provided in Fig. \ref{fig:cm_example}(c), where the same color labelling convention of \cite{Wu2018} is followed, with the green channel corresponding to the source mask, the red channel to the target mask and the blue channel to the background mask.
The final binary tampering mask for the image, where only the target region is highlighted (corresponding to the tampered part), is reported in Fig. \ref{fig:cm_example}(d).

\subsection{Problem formulation and rationale of the proposed solution}
\label{sec.formulation_rationale}

As we said, our goal is to devise a method for source-target disambiguation that exploits the {\em non-invertibility} of the copy-move process caused by interpolation.
To improve the effectiveness of the algorithm, we also exploit the possible presence of boundary artefacts in the target region (e.g. those due to blending), which are not present in the source.
In fact, even if copy-move tampering is carried out properly, subtle boundary artfacts and edge inconsistencies are often present and can be exploited for the disambiguation task. 

\begin{figure}[!h]
	\centering
	\includegraphics[width=0.99\linewidth]{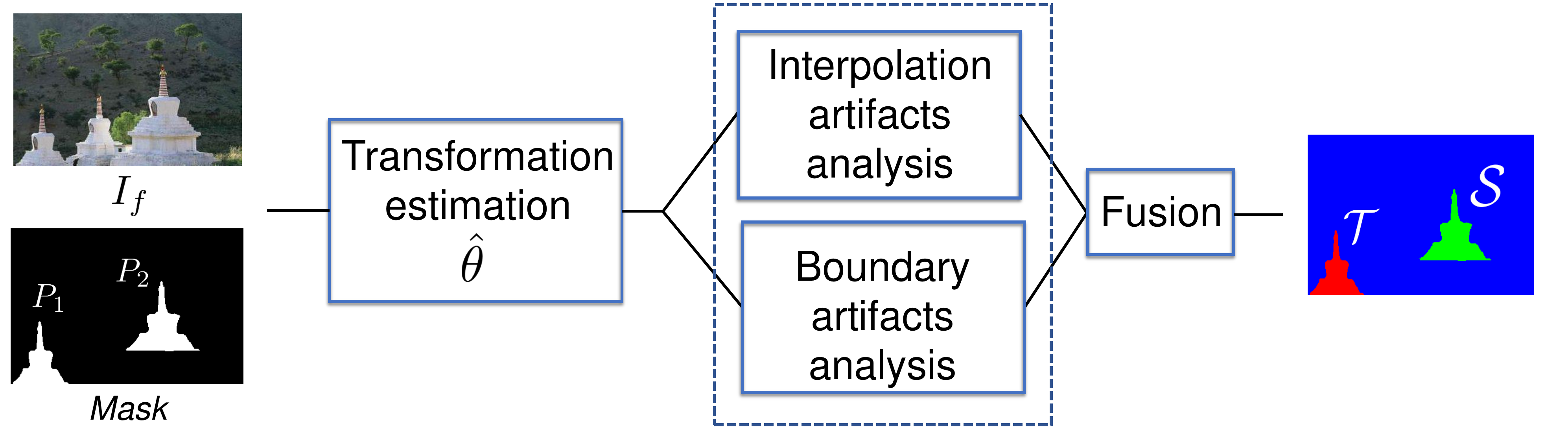}
    \caption{Scheme of the proposed CM disambiguation system.}
	\label{fig:GeneralScheme}
\end{figure}

The general scheme of the architecture we designed to solve the CM disambiguation problem is provided in Fig. \ref{fig:GeneralScheme}.
The input to the system are the forged image $I_f$, and the localization mask consisting of two separate regions. Note that we refer to the case of spatially separated regions for sake simplicity, however, the analysis is still valid for contiguous regions, assuming that the CM detection algorithm outputs two  distinct regions.

Let us first focus on the upper branch of Fig. \ref{fig:GeneralScheme}.
Given a pair of nearly duplicate regions, our first approach to disambiguate the source and target regions relies on the following observation:
if one tries to replicate the copy-move process starting from the source region, i.e. in the  {\em forward} direction, ideally, it is possible to re-obtain exactly the target region (in practice, the exact parameters of the transformation bringing $S$ into $T$ are not known exactly, so we will only obtain a very good approximation of $T$).
On the other hand, if one tries to mimic a copy-move process starting from the target region, i.e. in the {\em backward} direction, an exact copy (or even a good approximation) of the source region can not be obtained, due to the non-invertibility of the copy-move process.
In other words, when the target region $T$ is moved onto the source $S$, the approximated source region
differs significantly from $S$ due to the double interpolation process that the transformed region is subject to (from source to target, and then from target to source again), while no interpolation artefacts are present in the source region, thus making the approximation less close than in the opposite case, where both the target and the approximation of the target are subject to a similar (ideally the same) interpolation procedure.

Based on this idea, starting from the two regions and their approximated versions, the problem of disambiguating the source and target regions can be
formulated as the following {\em composite hypothesis test}. 
Let $P_1$  and $P_2$ denote the two nearly duplicate regions resulting from the binary localization map provided by the copy-move detector.
Then, the composite hypothesis test we have to solve must decide between the following cases\footnote{We use $\approx$ instead of $\equiv$, since the equivalence may not be satisfied due to the presence of post-processing.}:
%
\begin{itemize}
\label{HT}
\item $H_0$: ${P}_2 \approx \Psi_{\theta_0,k_0}({P}_1)$, i.e., $P_1\equiv S$ (and $P_2 \equiv T$).
\item $H_1$: ${P}_1 \approx \Psi_{\theta_1,k_1}({P}_2)$,  i.e., $ P_2 \equiv S$  (and $P_1\equiv T$)
\end{itemize}
%
%
where $\theta_0$ and $\theta_1$ are the parameters of the transformation bringing $P_1$ into $P_2$ and viceversa, and $k_0$ and $k_1$ are the interpolation kernel parameters. When hypothesis $H_0$ holds,
then  $\Psi_{\theta_0,k_0}$ corresponds to the transformation applied during the copy-move process, for some unknown parameter vector $\theta_0$ of the geometric transformation ${\cal H}_{\theta_0}$, and kernel $k_0$ of the interpolation.

To test the two hypotheses, we need to consider the transformation that moves ${P}_1$ to ${P}_2$,  and viceversa (i.e., the transformation that moves ${P}_2$ to ${P}_1$), and try to guess which of the two is the forward direction.
%
Therefore, as depicted in Fig. \ref{fig:GeneralScheme}, we should first estimate the parameters of the  transformation under both hypotheses
and then choose the direction for which the approximation obtained by means of the estimated transformation is the best one.
Formally, this is equivalent to solve the following generalized likelihood ratio test (GLRT):
\begin{equation}
%
\frac{\max_{\theta_0,k_0} Pr\{P_1 \approx \Psi_{\theta_0, k_0}({P}_2) | P_1,P_2, I_f\}}{\max_{\theta_1, k_1} Pr\{P_2 \approx \Psi_{\theta_1, k_1}({P}_1) | P_1,P_2, I_f\}} \underset{H_1}{\overset{H_0}{\gtrless}} 1.
\label{eq.GLRT}
\end{equation}
%
%
%
For simplicity,  the effects at the borders of the target region, due to possible local post-processing, are not taken into account  in the above formulation.

Since the interpolation method adopted for the copy move is unknown,  strictly speaking, it should be estimated. However, in our practical implementation, we have assumed that a bilinear interpolation is used, hence $k_0=k_1=k$, where $k$ is the bilinear kernel\footnote{Based on the experiments, the approach works with real-word copy-move datasets with possibly different interpolation methods; then, such a simplifying assumption is not too limiting.}. Then, we only estimate the parameters of the geometric transformations, that is, $\theta_0$ and $\theta_1$.

\CORR{Copy-move detection methods provide a binary localization
mask highlighting the regions interested by the copy move, yet only few of them
provide an estimate of the geometric transformation mapping
one region into the other (e.g., the keypoint-based detector in \cite{Amerini2013}, where an estimation of the transformation is provided  via the RANSAC algorithm \cite{fischler1981random}).
Therefore, in the first step of the disambiguation chain in Fig. \ref{fig:GeneralScheme}, we estimate the
geometric transformation bringing $P_1$ into $P_2$ (and viceversa).
%
Such estimation  can be performed in several ways leading to similar, yet not exactly identical results. The exact procedure adopted in our system is provided in the Appendix for sake of reproducibility.
}

\CORR{We let $\widetilde{P}_2 = \Psi_{\hat{\theta}, k}(P_1)$, i.e. the approximated $P_2$ region, where $\hat{\theta}$ is the estimated vector of the parameters of the transformation that moves  $P_1$ into $P_2$  (w.l.o.g.); similarly, the approximated  $P_1$ region is $\widetilde{P}_1 = \Psi_{\hat{\theta}',k}(P_2)$, where $H_{\hat{\theta}'} = {H_{\hat{\theta}}}^{-1}$ (in this way the transformation is estimated in one direction only).}

\begin{figure}[!h]
	\centering
	\includegraphics[width=0.99\linewidth]{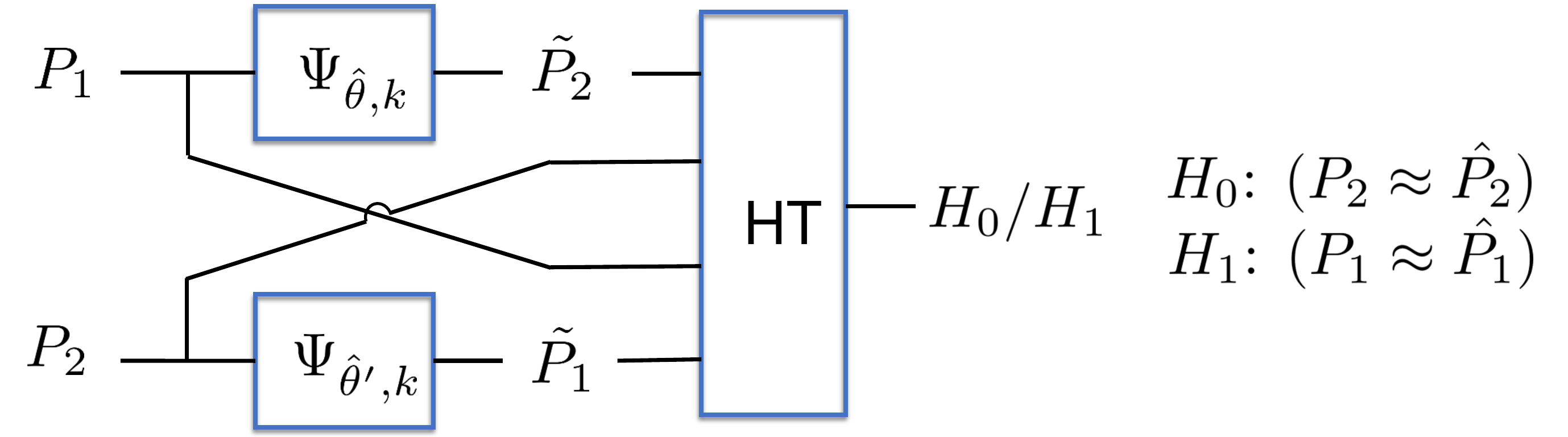}
    \caption{\BTT{Block diagram of the disambiguation test in \eqref{eq.GLRT}, where ${\theta}'$ is the vector of parameters such that $H_{\hat{\theta}'} = {H_{\hat{\theta}}}^{-1}$.}}
	\label{fig.blockSchemeHT}
\end{figure}

\BTT{A block diagram of the disambiguation test in \eqref{eq.GLRT},
fixing the notation, is provided in Fig. \ref{fig.blockSchemeHT}.}

When the interpolation artefacts are weak or not present at all, e.g., when the copy-move consists of a rigid translation of an integer number of pixels, we  have  $P_1 \approx \Psi_{\theta_0, k}({P}_2)$ and $P_2 \approx \Psi_{\theta_1, k}({P}_1)$, with ${H}_{\theta_1} = {H}_{\theta_0}^{-1}$ (the kernel $k$ is close to a delta function), and then we cannot make a reliable decision based on the test in \eqref{eq.GLRT}.
The bottom branch of the scheme in Fig. \ref{fig:GeneralScheme} is introduced to cope with these cases.
Such a branch exploits the possible presence of artefacts along the
boundaries of $P_1$ and $P_2$. For the target region, in fact, boundary artefacts are likely to be present given that the inner and outer parts of $T$ come from different parts of $I$. These artefacts are not expected to be present across the boundary of $S$. Therefore, the presence of such artefacts or other inconsistencies along the boundary of one region between $P_1$ and $P_2$ can be exploited to decide which of the two regions correspond to $S$ and which to $T$.
A further motivation for the inclusion of a branch dedicated to the presence of artefacts along region boundaries, is that the interpolation traces could be partially erased when a strong post-processing is applied globally, thus making it difficult to solve the disambiguation problem via the composite test formalized above. In these cases, the analysis of boundary inconsistencies can be useful.

Eventually, the result of the analysis of interpolation and boundary artefacts is fused (last block in Fig. \ref{fig:GeneralScheme}).

\section{Multi-Branch CNN Architecture}
\label{sec.multibrachCNN}

\begin{figure}[!h]
	\centering
	\includegraphics[width=0.85\linewidth]{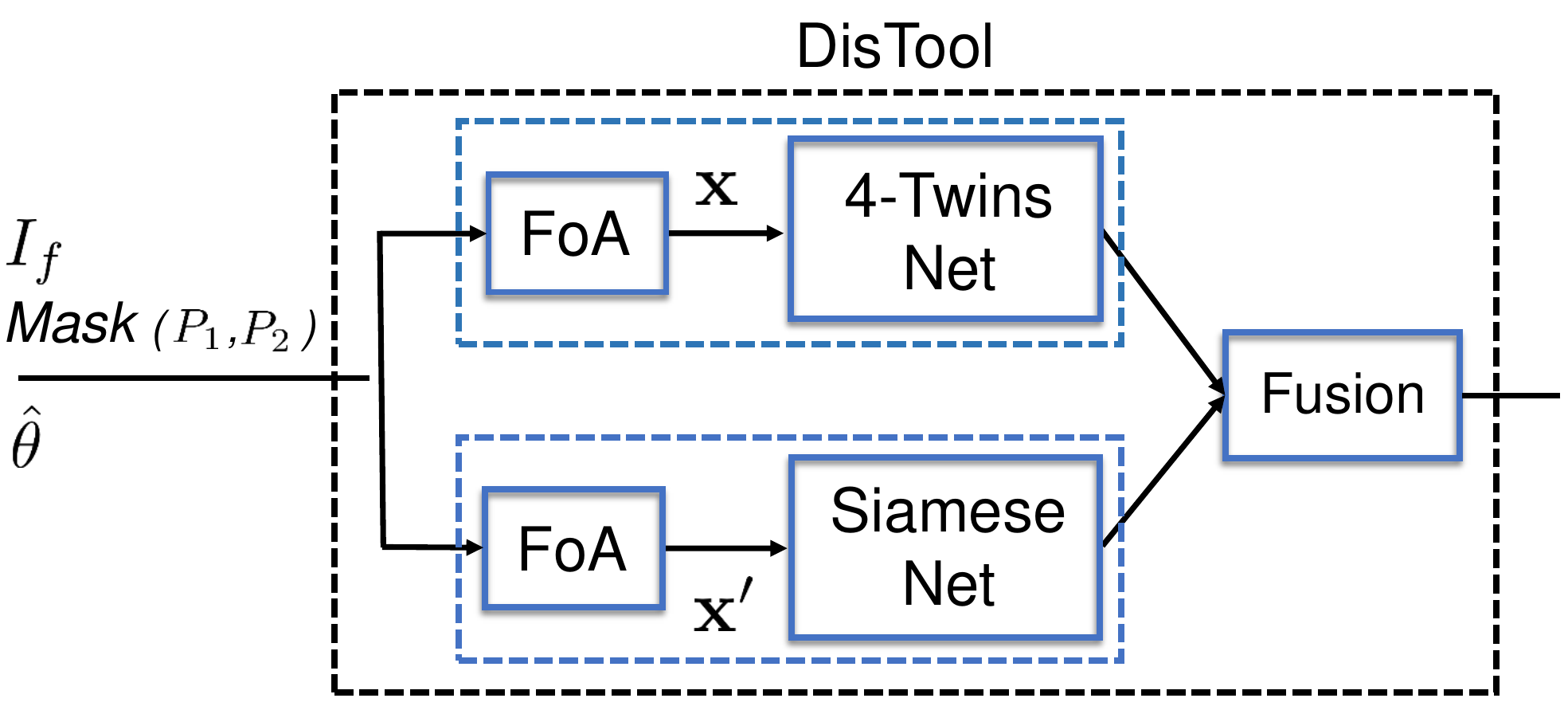}
    \caption{Block diagram of the proposed disambiguation method based on multiple-branch CNNs  (\textsf{DisTool}), implementing the scheme in Fig. \ref{fig:GeneralScheme}.}
	\label{fig:scheme}
\end{figure}

The core of the disambiguation system is represented by the blocks that analyze the interpolation artefacts and the boundary inconsistencies (see Fig. \ref{fig:GeneralScheme}). For their implementation, we designed two multiple-branch classifiers based on CNNs: a  network with 4 parallel branches, called {\ournet}, and a {\em Siamese} network \cite{Bromley1993}, named {\oursiam}. The 4-Twins network is in charge of analysing the interpolation artefacts, while the Siamese network is used to reveal boundary inconsistencies.
The outputs of the two networks are finally merged by a score-level fusion module.
\CORR{Other works in the forensic literature that resort to networks to learn inner traces and boundary artifacts for splicing and manipulation detection are \cite{salloum2018image,zhou2018generate}.}
A block diagram of the proposed architecture, hereafter referred to as \textsf{DisTool},
is shown in Fig. \ref{fig:scheme}.
A preliminary step is carried out before running the two networks to identify the input region, or Focus of Attention (FoA), of the networks.
Each FoA module takes as input the forged image $I_f$, the binary localization mask (i.e., the output mask of the CM detection algorithm) with the two separate regions $P_1$ and $P_2$, and the geometric transformations estimated as explained in the Appendix.
%
%


We observe that, while Siamese-like architectures have recently been used for addressing several multimedia forensic tasks, see for instance \cite{Cozzolino2018, Mayer2018, Huh2018}, we explicitly designed the 4-Twins architecture for our specific purpose, in order to facilitate the learning of the interpolation artefacts.
The motivation behind the use of this multi-branch architecture will be more clear in the sequel.

\subsection{{\ournet}}
\label{sec.4Twins}

The 4-Twins network takes as input the two pairs of regions $(P_1, \widetilde{P}_1)$ and $(P_2, \widetilde{P}_2)$ (the specific FoA
for {\ournet} is described in Section \ref{sec.FoA_4T})

\begin{figure}[!h]
	\centering
    \vspace{-0.4cm}
	\includegraphics[width=0.8\linewidth]{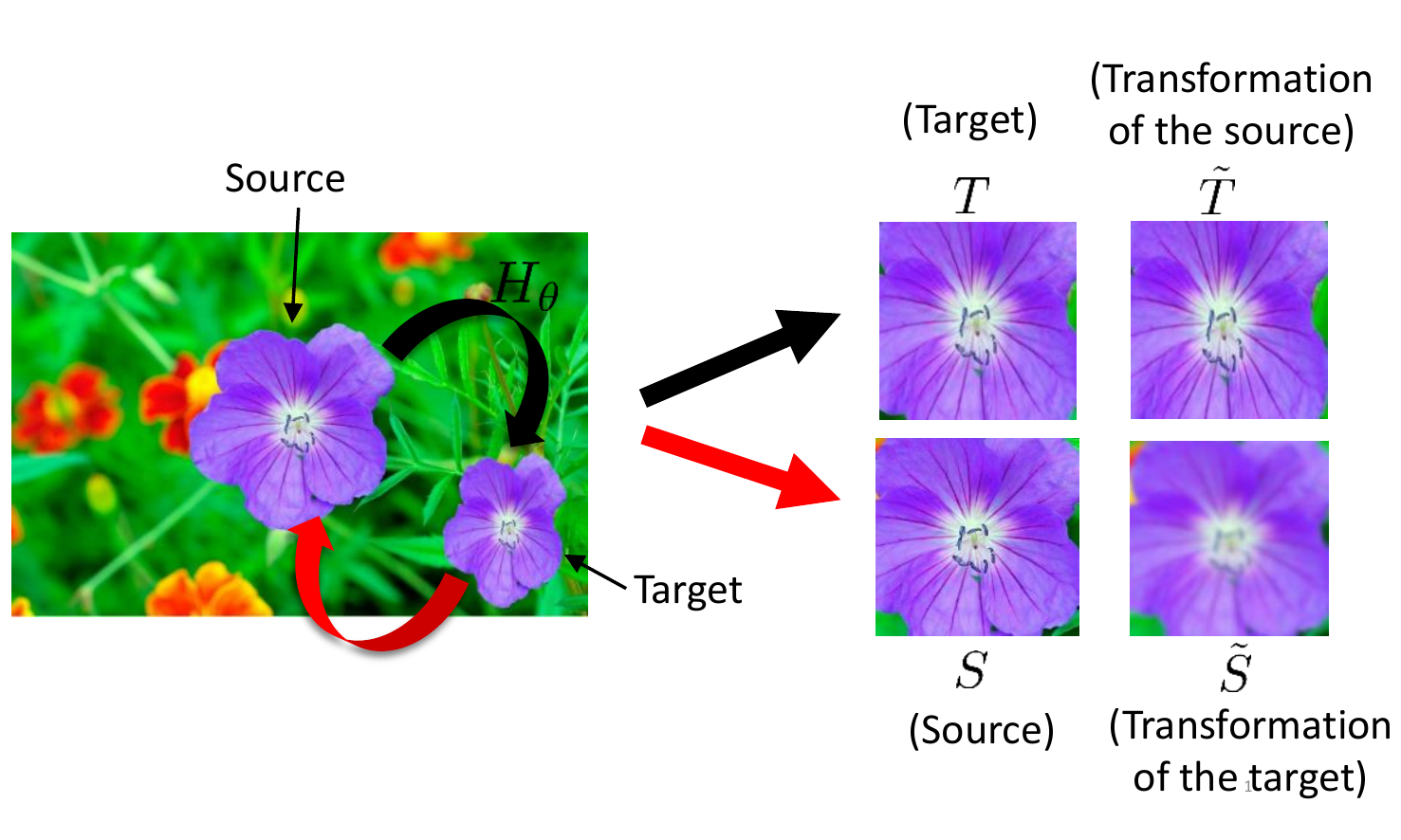}
    \includegraphics[width=0.8\linewidth]{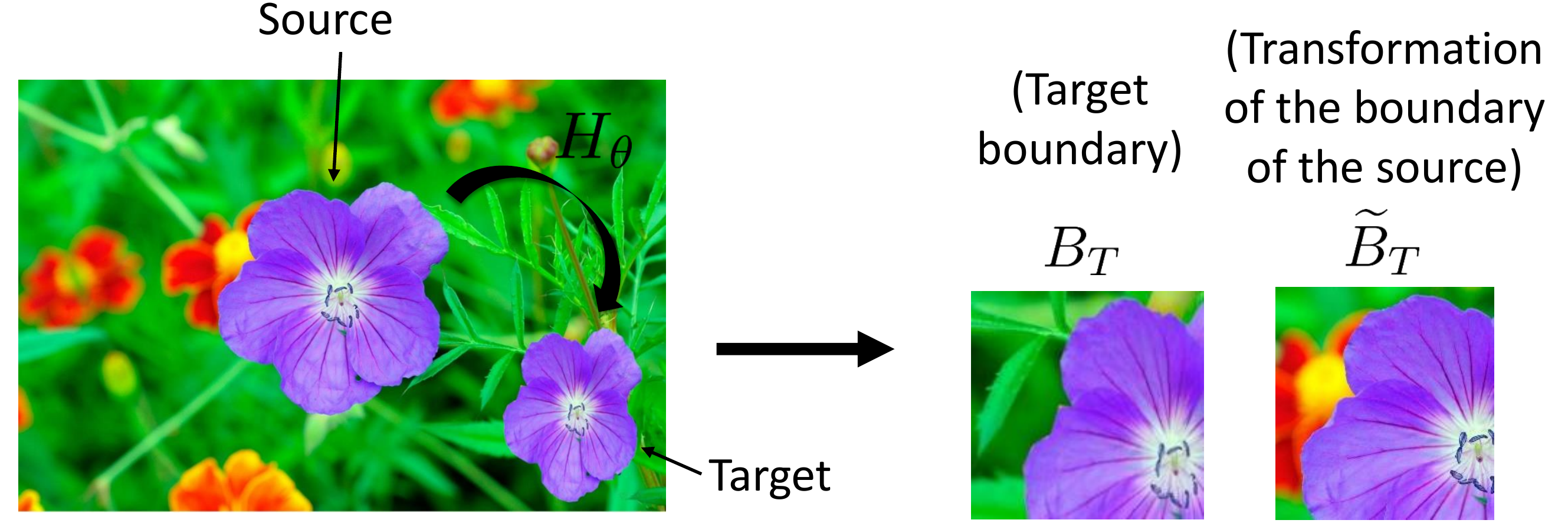}
    \caption{Illustrative example of the patches at the input of {\ournet} (upper row) and {\oursiam} (lower row).}
	\label{fig:inputs}
\end{figure}

Let
${\bf x}= [x_1, x_2, x_3, x_4] = [ (P_1, \widetilde{P}_1), (P_2, \widetilde{P}_2) ]$  be a vector with the pixels of the regions $P_1, \widetilde{P}_1, P_2,$ and $\widetilde{P}_2$, and let $y \in \{0,1\}$ indicate the identity of the source and target regions, namely $y = 0$ if ${\bf x}=[ (S, \widetilde{S}), (T, \widetilde{T}) ]$ (holding under hypothesis $H_0$), and  $y = 1$ if ${\bf x}=[(T, \widetilde{T}), (S, \widetilde{S})]$ (holding under hypothesis $H_1$).
An illustrative example of the patches at the input of {\ournet} is provided in Fig. \ref{fig:inputs} (upper row).

The decision is in favor of the hypothesis that maximizes the output score (softmax) function $f_{tw}$.
Therefore, if
\begin{align}
f_{tw}({\bf x}|y = 0)  >   f_{tw}({\bf x}| y = 1),
\end{align}
%
then $P_1$ is identified as the source region ($H_0$ holds);
viceversa, if the opposite inequality holds, $P_1$ is identified as the target ($H_1$ holds).



\CORR{Compared to a 2-branches architecture accepting only two inputs, e.g. $(P_1, \widetilde{P}_1)$ or $(P_2, \widetilde{P}_2)$,
an architecture that simultaneously takes all the 4 inputs allows to exploit the prior information available on the problem, that is the fact that if one direction is recognised as the forward direction, the opposite one has to be consistent with a backward direction  and viceversa\footnote{\CORR{This means that if the  patches in the first pair are similar to each other, than the patches in the second pair should contain some dissimilarities due to the wrong direction of the transformations - see discussion in Section \ref{sec.formulation_rationale}.}}. Based on the experiments we carried out, this way of exploiting the a-priori information gives an advantage to the 4-branch architecture.}

\CORR{From a more general perspective, one may wonder if using a simple similarity metric would be enough to distinguish between the two alternatives. For instance, we could use the MSE between the pairs and choose the transformation direction corresponding to the pair with a lower MSE. The performance achievable with such an approach, however, are not very good especially in the more realistic case where the localization masks are not ideal (as it is the case when the mask is estimated by a CM algorithm), the transformation is not estimated perfectly, and when the size of the copied region is small. Some results obtained by using the MSE-based disambiguator are reported in Section \ref{res.EtoE}.}

The architecture of {\ournet} and the details of the training procedure are described in the following.
Before that, we give the details of the FoA module.

\subsubsection{Focus of Attention (FoA)}
\label{sec.FoA_4T}

The two pairs of regions $(P_1, \widetilde{P}_1)$ and $(P_2, \widetilde{P}_2)$ can not be directly fed to {\ournet}.
%
The practical problem is that the source and target regions of a copy-move can be large and, moreover, their sizes vary from image to image.
In order to feed all the branches with patches of the same size, that we set to $64 \times 64 \times 3$,
the 4-dim input vector ${\bf x}$ of {\ournet} is built as follows (the first steps are common to {\oursiam}).
Given the two regions $P_1$ and $P_2$,
we fit a rectangular bounding box to each region. Let us denote the bounding box of $P_1$ as $P_1^b$. The bounding box will then contain the entire region $P_1$ (foreground) and some neighboring pixels belonging to $\bar{P}_1$ (background). In the same way, we build the rectangular patch $P_2^b$.
Then, we compute $\widetilde{P}_2^b = \Psi_{\hat{\theta},k}(P_1^b)$  and  $\widetilde{P}_1^b = \Psi_{\hat{\theta}', k}({P}_2^b)$, using bilinear interpolation.
%
In this way, we get the quadruple $[({P}_1^b, \widetilde{P}_1^b), ({P}_2^b, \widetilde{P}_2^b)]$.
To get the 4 inputs of {\ournet}, we crop the $64 \times 64$ central part of each region in the quadruple. Notice that, in this way, we are implicitly assuming that the bounding boxes of the source and target regions of the copy-move regions are always larger than  $64 \times 64$ (hence,  $64 \times 64$ is considered as minimum region size).
%
%

\begin{figure}
\centering
		\includegraphics[width=0.8\linewidth]{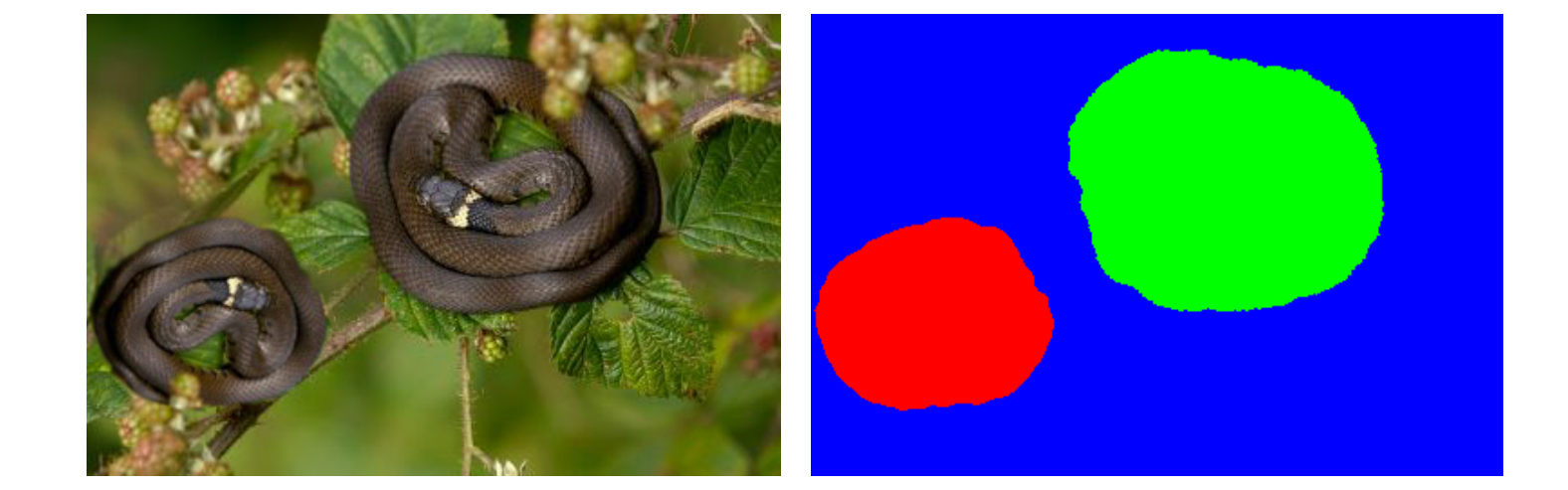}
		\caption{Forged image from CASIA \cite{Dong2013} (left) and its ground truth tampering map (right).}
\label{fig:exampleCASIA}
\end{figure}

\begin{figure}
\centering
		\includegraphics[width=\linewidth]{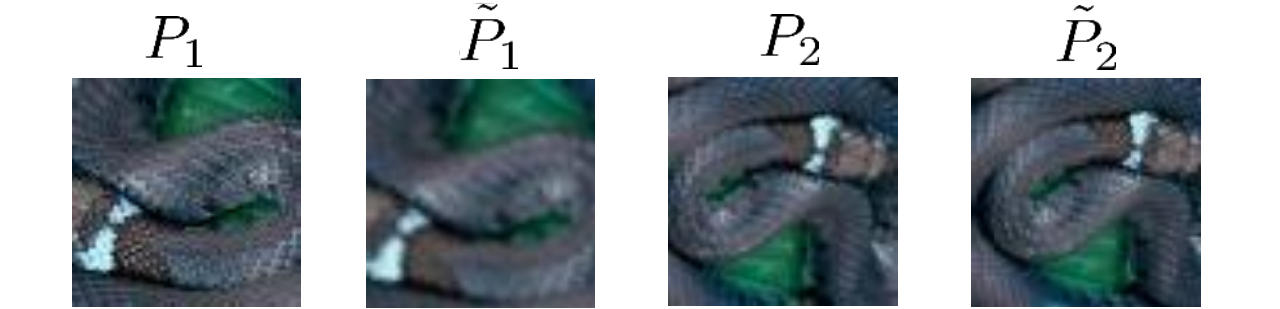}
		\caption{Input vector for {\ournet} for the image in Fig. \ref{fig:exampleCASIA} (central crop). }
	\label{fig:pair_extraction_4T}
\end{figure}

To avoid complicating  the notation, in the following we will not distinguish between regions and patches and continue to refer to $P_1$, $P_2$,
and $\widetilde{P}_1$, $\widetilde{P}_2$
to denote the inputs of {\ournet}.
An example of the patches forming the input vector ${\bf x}$ of {\ournet} is given in Fig. \ref{fig:pair_extraction_4T}  for the example in Fig. \ref{fig:exampleCASIA}.

\subsubsection{Network Architecture} 
\label{sec.4Twins_arch}

\begin{figure}[!t]
\centering
	\includegraphics[width=0.85\linewidth]{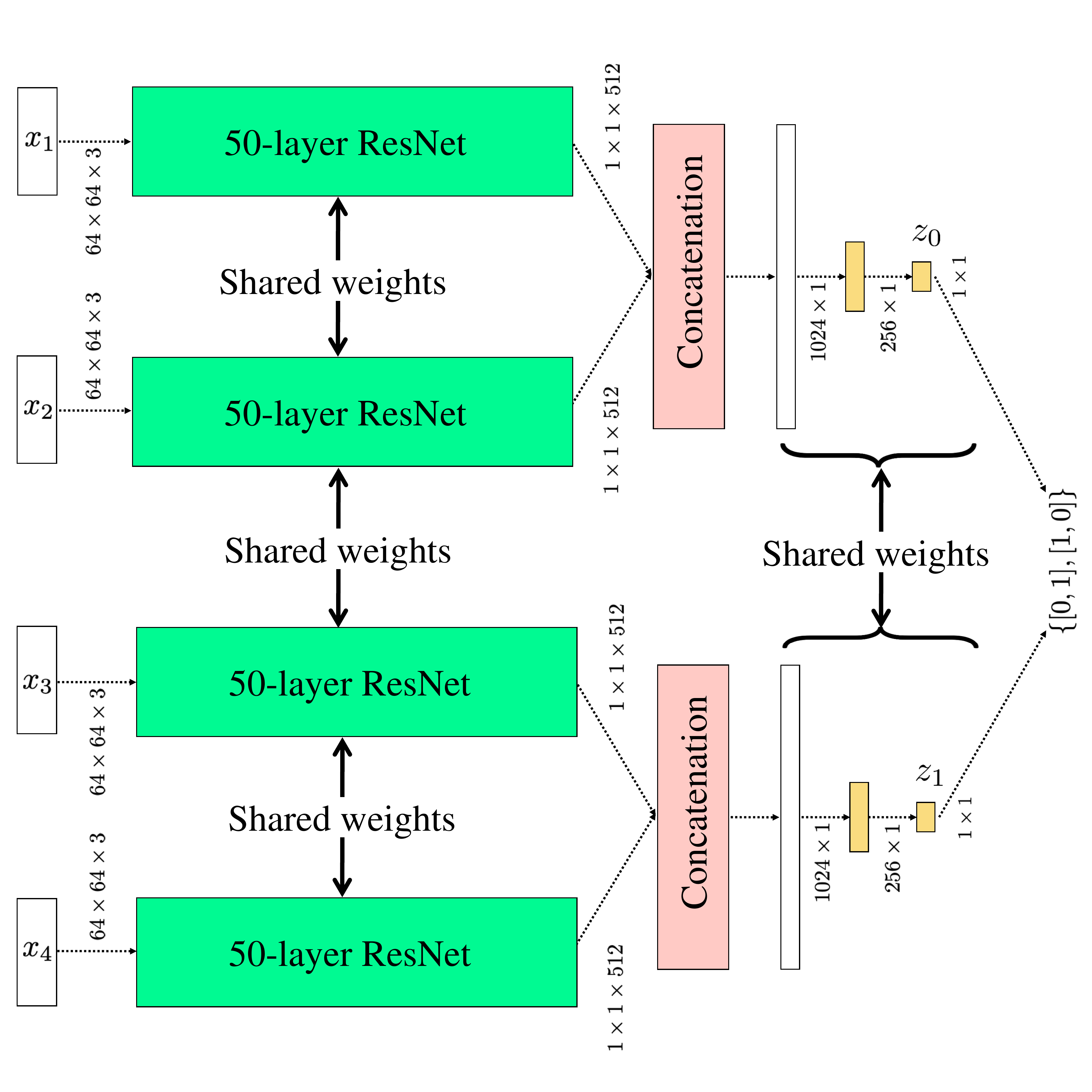}
	\caption{The proposed {\ournet}  architecture. The one-hot encoding of the predicted label is reported in output.}
	\label{fig:4_twin_net}
\end{figure}

The architecture of the {\ournet} is given in Fig. \ref{fig:4_twin_net}. It consists of four identical stacks of convolutional layers (i.e. all of them share the same weights), and two identical stacks of fully connected layers. The role of the stacked convolutional layers in each branch is to extract a $512$-dim feature vector from each input patch, of size $64 \times 64 \times 3$. We denote the stacked convolutional layers as  $\mathcal{F}(\cdot)$. The $512$-dim feature vectors from the first and second pairs of branches  are concatenated by means of a combination function $\mathcal{C}(\cdot, \cdot)$  in a 1024-dim vector, and then given as input to the fully connected layers.
%
%
The fully connected  layers return a  score (called \textit{logit}) which is later normalized into a probability value by means of softmax non-linear activation functions.
In summary, the 4-Twins architecture consists of two Siamese networks in parallel, sharing the weights of the convolutional layers and the fully connected layers.

For each Siamese network, we used exactly the same pipeline which has been successfully used as a matching model in computer vision \cite{Chopra2005,Koch2015}.
Each Siamese network has a single output neuron.
\CORR{Let $z_0, z_1$ denote the outputs (logits)
of the two Siamese network branches with inputs $(x_1,x_2)$ and $(x_3,x_4)$, respectively (see Figure \ref{fig:4_twin_net})}. The dependency between $z_0$ and $z_1$ is enforced by the following softmax operation:
%
\begin{equation}
\label{4Twins_output}
	f_{tw}({\bf x}|y=i) = \frac{e^{z_{i}}} {\underset{j \in \{0,1\}}{\sum}  e^{z_{j}}}, \quad i = 0,1.
\end{equation}
%
Given $M$ training examples  $\{{( {\bf x}^{(j)}, y^{(j)})}_{j \in [1,M]}\}$,
the {\ournet} is trained to minimize the empirical cross entropy loss function between input labels and predictions, that is:
%
\begin{equation}
\mathcal{L} = - \frac{1}{M} \sum_{j=1}^M \sum_{i \in \{0,1\}} \bigg( y_{i}^{(j)} \log f^{(j)}_{tw} \left({\bf x}|y=i \right)\bigg),
\end{equation}
where $[y_{0}^{(j)}, y_{1}^{(j)}]$ is the one-hot encoding of $y^{(j)}$ (the one-hot encoding of label 0 is the binary vector $[0,1]$, that of label $1$ is $[1,0]$).
\CORR{ From \eqref{4Twins_output}, we observe that,
$z_0$ will be large (and then $z_1$ will be small)  when $y = 0$ (i.e., under $H_0$), that is, when $P_1$ is the source, and small  ($z_1$ large)  when $y = 1$ (i.e., under $H_1$), that is, when $P_1$ is the target.}

In the following, we report the details of the feature extraction, combination and fully connected part of each Siamese branch.

%

\textbf{Feature extractor} $\mathcal{F}$.
%
%
%
We considered the 50-layers Residual Network (ResNet) in \cite{He2016}.
Such a deep architecture is well suited to learn complex pixel relationships \footnote{Based on preliminary tests that we carried out, shallow architectures do not permit to achieve high accuracy for our task.}. We refer to \cite{He2016} for a detailed description of this network.
The only change we made compared to  \cite{He2016} is the output size, which is set to $512$ instead of $1000$.
Then, in our architecture, we considered 4 identical branches of 50-layers ResNet ($\mathcal{F}$), with shared weights, for the convolutional part.

\textbf{Feature combiner} $\mathcal{C}$.
Before feeding the fully-connected layers, we need to fuse the feature vectors produced by the two Siamese branches $\mathcal{F}$. Some popular choices for doing so are: the point-wise absolute difference \cite{Koch2015}, the square Euclidean distance \cite{Cozzolino2018}, and the concatenation \cite{Huh2018}.
We chose to implement the combination by means of a concatenation as done in \cite{Huh2018}.

\textbf{Fully connected part}. We considered 2 fully connected layers with input and output sizes respectively equal to  $1024$ and $256$, and, $256$ and $1$. The final soft output of the two fully connected branches are combined by means of a softmax layer, as detailed in the previous section.\\

\subsubsection{Training strategy}
\label{sec.tacticsTR-4T}

In this section, we describe the strategies that we followed to feed the data to {\ournet}  during training. During our experiments we found that such strategies are critical to the success of the {\ournet}.

The network is trained with both positive ($H_0$) and negative ($H_1$) examples, in equal percentage; then, the trained model minimizes the overall error probability over the training set.
To force the network to learn the interpolation artefacts,
the source and target regions  of the forged images used for training are purposely built so that they are always much larger than  $64\times 64$ (see Section \ref{sec:exp_syn_db} for the details of the  dataset creation process). In this way, the  $64\times 64$ input patches obtained by cropping the central part of the regions
contain only foreground pixels.
Training {\ournet} is performed knowing the ground truth localization mask and the exact geometric transformations between $S$ and $T$, that is, the forward and backward transformation $\mathcal{H}_{\theta}$.
Then, the  approximated regions  $\widetilde{P}_1^b$ and $\widetilde{P}_2^b$ are derived by considering the true transformation matrix  ${H}_{\theta}$.
A small random perturbation is applied in order to mimic a practical scenario in which the transformation estimation is not perfect.
Specifically, the true angle is perturbed by a random quantity in $[-\ang{5},\ang{5}]$ (with step $\ang{1}$), and the true resizing factor is randomly distorted by a value in  $[-0.1,0.1]$ (with quantization step 0.01).


Due to feature concatenation, the network is sensitive to the order of the inputs in each pair, that is in $(x_1, x_2)$ and $(x_3, x_4)$. Let us assume that the first input corresponds to the original (source or target) patch and the second input to the transformed patch. In principle, switching between $x_1$ and $x_2$, as well as between $x_3$ and $x_4$, should leave the predictions  unchanged. To enforce this property,  we randomly shuffle $(x_1, x_2)$ and $(x_3, x_4)$ during training so that {\ournet} does not learn the order of the inputs. Under $y = 0$, this corresponds to consider not only the pair
${\bf x} = [(S, \widetilde{S}), (T, \widetilde{T})]$,  but also the pairs ${\bf x} = [(\widetilde{S}, S), (T, \widetilde{T})]$, ${\bf x} = [(\widetilde{S}, S), (\widetilde{T}, T)]$, and ${\bf x}= [(S, \widetilde{S}), (\widetilde{T}, T)]$. A similar strategy is applied under $y=1$.
Moreover, since the 4 branches of the convolutional layers are forced to be identical,  each of them is
fed with samples from all the categories during training, that is \{$S$, $\widetilde{S}$, $T$, $\widetilde{T}$\},  so to avoid any bias.

Batch Normalization (BN) \cite{Ioffe2015} is performed after each layer in the feature extraction part $\mathcal{F}$, by normalizing the layer outputs so that they have zero-mean and unit-variance. Normalization is done by accumulating means and standard deviations on mini-batches. Given that in  {\ournet},  the data flows through four branches, this procedure needs care: in particular, in order to avoid biasing the accumulated means and standard deviations, we ensure that, within each mini-batch, each of the four branches is fed with all four categories \{$S$, $\widetilde{S}$, $T$, $\widetilde{T}$\}.
Then, the statistics are accumulated on one branch only and broadcasted to the other branches in order to make the four $\mathcal{F}$s identical.

%

\subsection{{\oursiam}}
\label{sec:siamese_net}


As we said, the goal of the {\oursiam}
is to detect boundary inconsistencies.
The choice of this structure was based on the following observation.
When $P_1$ is the source ($H_0$),  we expect that the pixels across the boundary of $P_1$ and the complementary region $\bar{P}_1$  do not present significant  inconsistencies, while, when $P_1$ is the target ($H_1$), the presence of inconsistencies along the boundary between  $P_1$ and $\bar{P}_1$ is more likely.
%
%
Let $B_{1}$ (res. $B_{2}$) denote an image region that includes $P_1$ (res. $P_2$) and some outer pixels of $P_1$ (res. $P_2$) , i.e., part of the complementary region $\bar{P}_1$ (res. $\bar{P}_2$).
Similarly, $B_S$ (res. $B_T$) denotes an
image region that includes $S$ (res. $T$) and some outer pixels of $S$ (res. $T$).
%
Regions $B_{1}$ and $B_{2}$ have the same (or very similar) content inside the inner region and a different content in the outer part. So we would like that the network learns to focus on the relationship between the inner and outer region. i.e. to focus on the values of the pixels across the boundary of the copied part.
%
%

The transformation that maps one region into the other, e.g. $P_1$  into  $P_2$, also maps (at least approximately because of the interpolation) the boundary of $P_1$ into that of $P_2$.
Therefore, the Siamese network is fed
with input pairs  $(B_{1}, \widetilde{{B}}_{1})$ (or, similarly, $(B_{2}, \widetilde{{B}}_{2})$) where
$\widetilde{B}_{1}$ is obtained by remapping  $B_{2}$ according to the geometric transformation that maps  $P_2$ into  $P_1$,
that is, $\widetilde{{B}}_{1} = \Psi_{\hat{\theta},k} ({B}_{2})$. 
%
The details about the exact way whereby the regions $B_1$ and $B_2$ are built, pertaining to the FoA block preceding  {\oursiam}, are described in Section \ref{sec.FoA_Siam}.
Let ${\bf x}'= [x_1', x_2'] = [ B_{1}, \widetilde{B}_{1}]$ and $y \in \{0, 1\}$.
The relative position of the patches in the pair determines the value of $y$:
if ${\bf x}'=[B_S, \widetilde{B}_S]$,  that is $P_1 \equiv S$ (hypothesis $H_0$), then $y = 0$; if instead ${\bf x}'=[B_T, \widetilde{B}_T]$, that is $P_2 \equiv S$ (hypothesis $H_1$),  then $y = 1$. 

An illustrative example of input pairs feeding the {\oursiam} is shown in Fig. \ref{fig:inputs} (lower row).

The decision is in favor of the hypothesis that maximizes the output soft function $f_{si}(\cdot)$. Therefore, the condition
\begin{align}
f_{si}([ B_1, \widetilde{B}_1]| y=0)  > f_{si}([B_1, \widetilde{B}_1]| y=1),
\end{align}
indicates that $P_1$ is the source ($H_0$ holds), while the opposite inequality indicates that $P_2$ is the source ($H_1$ holds).

We notice that the use of an architecture with 4 branches, like the {\ournet}, is not necessary in this case. In fact, regardless of the direction of the transformation, one  region between $B_1$ and $\widetilde{B}_1$, will exhibit  inconsistencies between the pixels inside and those outside the boundary, while the other will not. In a similar way, one between $B_2$ and $\widetilde{B}_2$ will contain inconsistencies across the boundary, while the other will not.

\BTT{With respect to using a single-branch architecture, each time taking only one between $B_1$ and $\widetilde{B}_1$ (or $B_2$ and $\widetilde{B}_2$) as input, showing simultaneously both regions to the network permits to learn relevant features more easily. The advantage of the Siamese network architecture over a single-branch CNN was confirmed by some preliminary experiments we carried out.}

The FoA, the architecture of {\oursiam} and  the details of the training procedure are described in the following.

\subsubsection{Focus of Attention (FoA)}
\label{sec.FoA_Siam}

To get the input pair ${\bf x}'$ for  {\oursiam}, we start with one of the pairs of bounding box regions $({P}_1^b, \widetilde{P}_1^b)$ and  $({P}_2^b, \widetilde{P}_2^b)$, obtained as described in Section \ref{sec.FoA_4T}.
In order to increase the chance of capturing a good extent of boundary regions,  each bounding box region is cropped at the 4 corners, i.e., top left, top right, bottom left, bottom right to get the $64\times 64$ input patches. All the resulting 4 input pairs are tested and the most confident prediction score is selected for the final decision.




To avoid complicating  the notation,  we continue to use $B_1$, $B_2$  and $\widetilde{B}_1$, $\widetilde{B}_2$, to denote the inputs of  {\oursiam}.

\begin{figure}[!h]
\centering
 \includegraphics[width=0.9\linewidth]{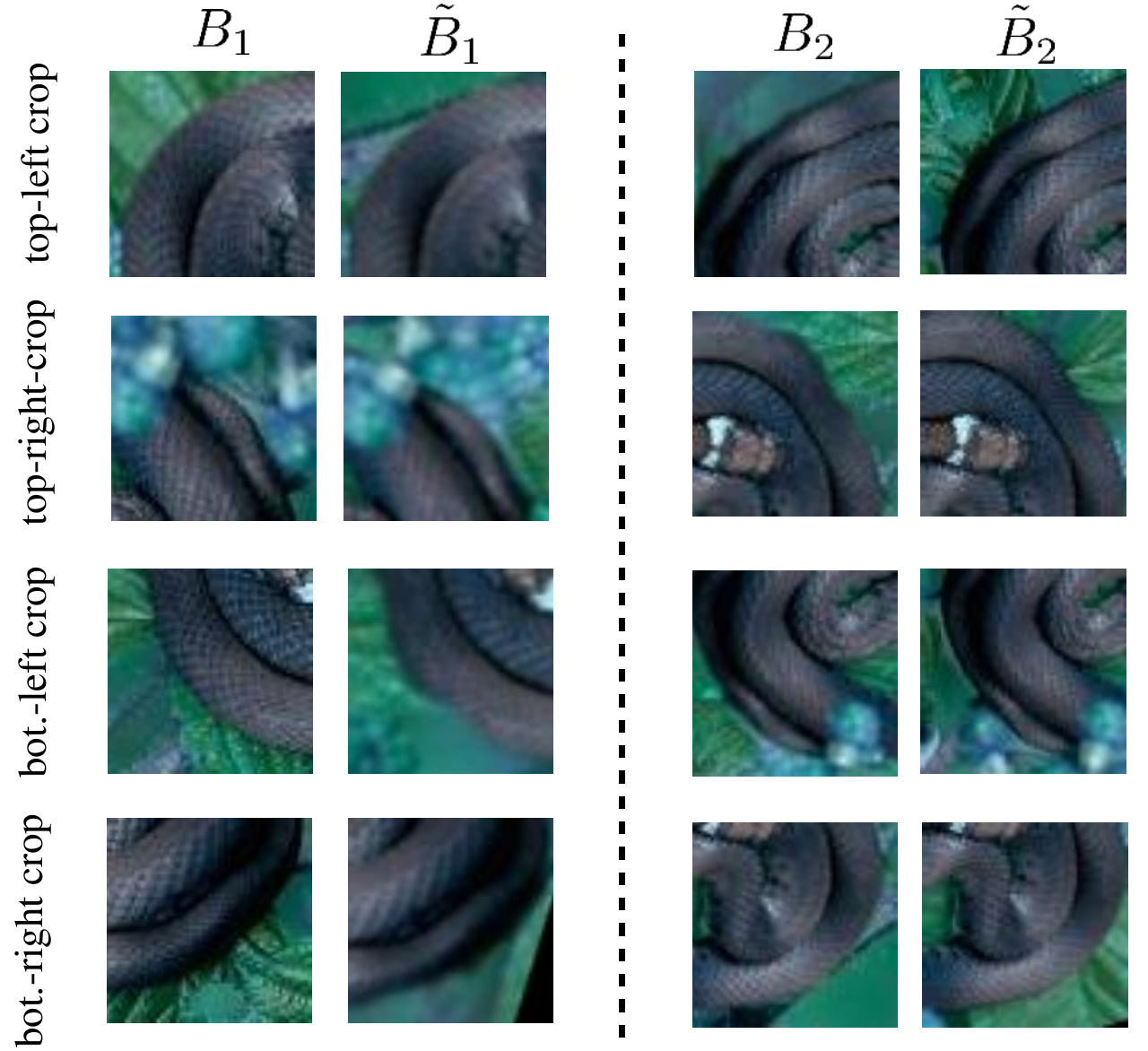}
 \caption{Examples of the input pairs for {\oursiam}, for the image in Fig. \ref{fig:exampleCASIA}.
  $y = 0$ for the examples of input pairs in the left  (since $B_1 = S$), while for those on the right $y = 1$  (since $B_2 = T$).}
\label{fig:pair_extraction_Siamese}
\end{figure}

The possible test input vectors for  {\oursiam}, for the example in Fig. \ref{fig:exampleCASIA}, are provided in Fig. \ref{fig:pair_extraction_Siamese}.
%

\subsubsection{Network architecture}
\label{sec.Siamese_arch}

The architecture of {\oursiam}
corresponds to the one forming the branches of {\ournet}.
Let $z$ be the output (logit) of the Siamese neural network, the soft (probabilistic) score $f_{si}$ is computed through a sigmoid activation:
\begin{equation}
	f_{si} \left ({\bf x} | y=1 \right) = \frac{1}{1 + e^{-z}} \text{.}
\end{equation}
Given $M$ training examples  $\{\left( {\bf x}^{(j)}, y^{(j)} \right)_{j \in [1,M]}\}$,
the {\oursiam} is trained to minimize the empirical cross entropy loss $\mathcal{L}_{si}$ between the predictions $f^{(j)}_{si}$ and the input labels $y^{(j)}$.\\
%


\subsubsection{Training strategy}
\label{sec.tacticsTR-Siam}

%

To force the network to look at boundary inconsistencies, we trained {\oursiam} by considering only copy moves obtained by rigid translations, so that: i) no interpolation artefacts are present (the copy moved part is identical to the source region), ii)
the boundaries of the two regions match perfectly. The input pair used during training then corresponds to ${\bf x}'=  [ B_{S}, B_T]$ and ${\bf x}'=  [ B_{T}, B_S]$.

In order to avoid undesired biases, the network is fed with inputs of the form ${\bf x}' = [B_S, B_T]$ and  ${\bf x}' = [B_T, B_S]$ in a similar percentage. Note that in this case, switching $x_1'$ and $x_2'$ is accompanied by  label switching: in fact,
according to the way we trained the network, the output of {\oursiam} depends on the relative position of the patch containing the source boundary and the target boundary.
Therefore, if ${\bf x}' = [B_S, B_T]$, then $y=0$, whereas if  ${\bf x} = [B_T, B_S]$, then $y =1$.\\


\BTT{
Before concluding this section, we pause discuss the relationships with the use of the Siamese network performed in other works for forensic tasks, e.g., \cite{Cozzolino2018, Mayer2018, Huh2018} for camera fingerprint and splicing detection.}

\BTT{We first observe that the rationale behind the use of the Siamese network has some similarity.
In \cite{Cozzolino2018, Mayer2018, Huh2018}, the Siamese network is designed and trained to learn inconsistencies between background and foreground and then is fed with background patches and tampered patches in input.
In the copy-move application considered in this work, we expect that the background (and then source region) and the tampered area (target region) are consistent, given that they come from the same image (we are focusing on the case of a rigid copy-move, when the interpolation traces can not be used to distinguish between source and target regions, since this is the case the Siamese network is mainly designed for).
On the contrary, inconsistencies are expected to appear across the boundaries of the target and source regions (precisely, artefacts are expected to be present along the boundaries of the target region, and not along the boundaries of the source region), motivating why patches that contain both part of the inner and outer regions are used to feed the network.}

\BTT{The main difference in the usage of the Siamese network we did in our case, with respect to the approaches in \cite{Cozzolino2018,Mayer2018,Huh2018}, is the following:  while in those cases the Siamese network is designed and trained in such a way  that it outputs a decision on the similarity or dissimilarity (i.e., on the consistency or not) of the input test patches, in our case, the test input pair always consists of a target ($B_T$) and source ($B_S$) boundary patch, the difference being in the order (the Siamese network is trained in such a way that it output 0 if the patches in the test pair are in the $(B_S,B_T)$ order, 1 for the $(B_T,B_S)$ order). Such behavior can be obtained through the use of the more general concatenation layer for feature combination, instead of a distance layer which is typically considered in the other applications, and yields to a different interpretation of the output score.}

\subsection{Fusion module}
\label{sec.fusion}

The output scores
$f_{tw}(\cdot)$ and $f_{si}(\cdot)$ provided by {\ournet} and {\oursiam} are fused by means of a simple fusion module, as illustrated in Fig. \ref{fig:scheme}.
Score-level fusion is performed by assigning a reliability to  the output of {\ournet} and {\oursiam}, based on the knowledge we have about the performance of the two networks  under various settings.
More specifically, the two  scores $f_{tw}(\cdot)$ and $f_{si}(\cdot)$ are weighted based on the (real or estimated) transformation mapping $P_1$ into $P_2$. Let us denote with $w_{tr}(\mathcal{H}_\theta)$ the weight assigned to the {\ournet} score when the estimated transformation  is $\mathcal{H}_\theta$ ($w_{tr} \in [0,1]$), and with $w_{si}(\mathcal{H}_\theta)$ the weight assigned to the output of {\oursiam}, where $w_{tr}(\mathcal{H}_\theta) + w_{si}(\mathcal{H}_\theta) = 1$.

We anticipate that, based on our tests, {\ournet} \CORR{works very well (with almost perfect performance)} when the transformation can be estimated with sufficient accuracy and relatively strong interpolation artefacts are present in the image, while it is less reliable in the other cases,
that is, basically, when the copy-move is close to a rigid translation. When the transformation is a rigid translation, in fact, the two input pairs of {\ournet} are identical.
\CORR{On the other hand, {\oursiam} achieves very good performance when a close-to-rigid translation is applied. Differently from the 4-Twins,  {\oursiam} is not specialized for the rigid translation case and its performance in the case of general transformations are also good.
}
%
 \CORR{With the above ideas in mind, in our experiments we considered the following  assignment for the weights:}
\begin{align}
\label{fusion_w}
w_{tr}(\mathcal{H}_\theta) =  \left\{\begin{array}{ll}
c &  \mbox{if ($|\alpha| > 15$ $\vee$ $|f_x - 1| > 0.1$ } \\
&  \mbox{\hspace{1.2cm} $\vee$ $|f_y - 1| > 0.1$)}\\
1-c & \mbox{otherwise}\end{array}\right.,
\end{align}
where $c$ is a constant larger than 0.5 ($\alpha$ denotes the rotation angle and $f_x$ and $f_t$ the scaling factors). As to $w_{si}$, we obviously have $w_{si} = 1 - w_{tr}$.


An alternative solution could be to choose one of the two networks before actually applying them, e.g. based on the estimated transformation  (network selection scenario).
However, fusing the outputs of both networks permits to get an advantage when  a choice between one of the two architectures cannot be properly made. As we will see in the experimental section, this is the case, for instance, when heavy local post-processing is applied to the boundary of the target region (in which case the {\oursiam} loses accuracy, while {\ournet} is more robust and still works well), or in the presence of global post processing, e.g. JPEG compression (in which case the performance of {\ournet} are heavily impaired). In these cases, fusing the outputs of both networks allows to get  better performance.

\section{Experimental Methodology and Settings}
\label{sec.methods}

In this section, we first describe the procedure that we followed to generate the synthetic datasets  used for training and validating
{\ournet} and {\oursiam}. Then, we present the datasets (both synthetic and real) used for testing. Finally, we describe the scenarios considered in our tests.

\subsection{Synthetic Dataset Creation}
\label{sec:exp_syn_db}


To train the multi-branch CNN, and in particular the {\ournet}, a large amount of labeled data is needed, i.e. many $({\bf x}, y)$ samples. Therefore, a large amount of copy-move forgeries with ground-truth mask and labeled source and target regions is required.
In \cite{Wu2018}, a dataset with $10^5$ copy-move forged images has been built and made publicly available.
This dataset, however, is too small for our goal.
To avoid the risk of overfitting, we synthesized ourselves a large-scale synthetic dataset of copy-move forged images
by considering several geometrical transformations and post-processing,  starting from pristine images of different datasets\footnote{The code used for
the creation of the dataset is made publicly available at \url{https://github.com/andreacos/MultiBranch_CNNCopyMove_Disambiguation}, for reproducibility. A python implementation of \textsf{DisTool}, and the trained models used for the tests, are also provided at the same link.}.
Specifically, a dataset with $9\times 10^5$ forged images, hereafter denoted as SYN-Tr was generated for training (and validation).  We also built a smaller set  of  $3\times 10^3$ forged images
for testing (namely SYN-Ts), as detailed in Section \ref{eval_datasets}.

The creation of SYN-Tr (and SYN-Ts) involves three steps.
%

\textbf{Background preparation}.
%
We first selected a pool of pristine images from several datasets. Specifically, approximately $28,000$ images (both in raw and JPEG formats) were taken from the RAISE\_2k \cite{Dang-Nguyen2015}, DRESDEN \cite{Gloe2010} and VISION \cite{Shullani2017} datasets to build SYN-Tr,
in similar proportions.
For SYN-Ts, we took $500$ images from a personal camera Canon 600D (250 raw and 250 JPEG
images, compressed using default camera settings.
%
For each image, we generated multiple forged instances
(as detailed below) by randomly cropping portions of size $1024 \times 1024$. The images having minimum dimension smaller than $1024$ were skipped.

\textbf{Source selection}. Each $1024 \times 1024$ image is split into four subregions or quadrants. The source region is obtained by considering one of these quadrants and generating a convex polygon (from a subset of $20$ random vertices, selected in such a way that they form a convex hull) within a bounding box of sizes $170 \times 170$, randomly located within the selected quadrant.
The pixels inside the convex polygon belong to the source region and then constitute the region $S$. 

\textbf{Target creation}. The target region is obtained from the source by means of a similarity transformation. In particular, we considered rotation, resizing, and a composition of them (i.e., rotation followed by resizing, and resizing followed by rotation). Rotation angles were randomly picked in the range $[\ang{2},\ang{180}]$, with a sampling step of $\ang{2}$, while horizontal and vertical resizing factors were randomly picked in $[0.5,2.0]$, with sampling step $0.01$.
%
%
The geometrically reshaped region is copy-pasted in the center of one of the three remaining quadrants, thus obtaining the target region $T$. With regard to the interpolation method, we used a bilinear interpolation. To improve the quality of the forged images making them more realistic,
\CORR{starting from the disambiguation map, we blurred the boundary of the target region by applying the following steps: i) detection of an edge enhanced mask, that is obtained by first enhancing the edges of the target region via a high-pass filter ($5 \times 5$) of the target mask to get the edge mask, then performing binary dilation for several iterations to emphasize the edges (the number of iterations is empirically set to 5), i.e., enlarge their tickness, thus getting the edge enhanced mask};\footnote{\BTT{Notice that the edges of the target region could be obtained from the knowledge of the polygon defining the source region and by the knowledge of the copy-move transformation, hence, strictly speaking, the edge detection step could be avoided.}}
ii) application of an average filter to the image, with a size randomly selected in $\{3 \times 3, 5 \times 5, 7 \times 7, 9 \times 9, 11 \times 11 \}$, in the positions identified by the edge enhanced mask.
%
%
%
Eventually, to mimic a real scenario, we applied global post-processing with probability $0.5$.
The post-processing types and the corresponding parameters are detailed in Table \ref{table:pp_details}, along with their selection probability.

Another dataset, named SYN-Tr-Rigid, was generated to train the {\oursiam}, by starting from the same pool of images, but considering only rigid translations. Source selection has been done within smaller bounding boxes of size $74 \times 74$ such that the boundary can be easily captured during the patch extraction process. Global post-processing is finally applied (with probability 0.5) similarly as before.
With regard to the test set SYN-Ts, for each kind of transformation (H), 1000 forged images
were generated, 500 with post-processing (PP)  - as described
in Table \ref{table:pp_details} - and 500 without postprocessing. In the following, we
denote with SYN-Ts-H and SYN-Ts-H-PP the datasets of test forged images
generated using transformation H, respectively without and with post-processing (PP). H can be a rigid translation (Rigid), rotation and translation (Rot), resizing and translation (Res).

\subsection{Evaluation Datasets} 
\label{eval_datasets}
%


\begin{figure}[t]
    \includegraphics[width=\columnwidth]{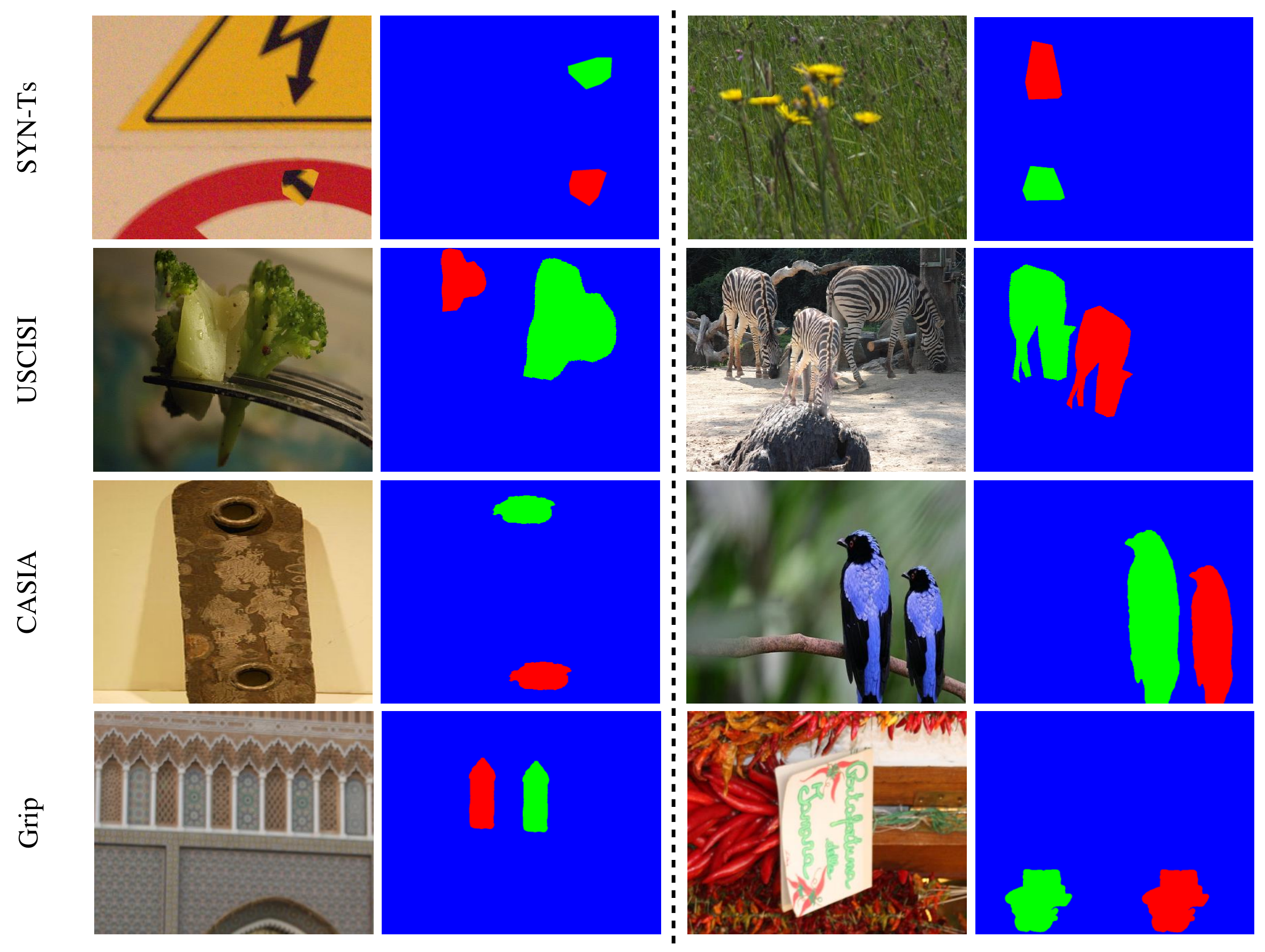}
	\caption{Examples of forged images from the 4 datasets used in our experiments. Red: target, green: source, blue: background.} 
	\label{fig:ex_db}
\end{figure}
We assessed the performance of our system on the datasets reported below, all providing ground truth mask and source-target labels for the copy-move forgeries.

	\begin{itemize}
		\item SYN-Ts. As detailed in Section \ref{sec:exp_syn_db}, this dataset contains $3\times 10^3$ test forged images.
		\item USCISI  \cite{Wu2018}. A synthetic dataset, consisting of
 $10^5$ images, that were used for training and testing BusterNet (in 9 to 1 proportion).
  All the images are taken from SUN2012 dataset \cite{Xiao2010} and Microsoft COCO \cite{Lin2014} that provide the object segmentation mask. Objects are copy-moved  by means of geometrical transformations (see \cite{Wu2018} for more details). For our tests, we used all the $10^4$ test images.
		\item CASIA \cite{Dong2013}. CASIA\footnote{http://forensics.idealtest.org/casiav2.} is the largest publicly available benchmarking dataset for image forgery detection. A subset of $1313$ copy-move forged images was manually selected, out of all the $5123$ tampered ones, by the authors of \cite{Wu2018}, to build this dataset of copy-moves, made available online. Source and target regions were labeled by comparing the tampered and the pristine images.
		\item Grip \cite{Cozzolino2014}. This dataset consists of $80$ images tampered with rigid copy-moves.
Two post-processing, i.e. local noise addition and global JPEG compression, were applied to these images, with different parameters, using the software in \cite{RiessTIFS}, thus producing several categories of copy-move forgeries.
We manually annotated the source and target regions of all the forged images by looking at the information on the top-left coordinates of the source and target regions provided by the software. Even if rather small, this dataset is useful to test the performance in the case of rigid copy-move.
	\end{itemize}
Some examples of copy-move forgeries from the four datasets are depicted in Fig. \ref{fig:ex_db}.

\begin{table}[!h]
\caption{Post-processing and corresponding probabilities.}
	\begin{tabular}{| L{1.8cm} | L{4.8cm} | L{0.8cm} |} \hline
		Processing type & {\centering Processing parameters} & Prob. \\
		\hline
		\hline
		Identity & -- & 0.5 \\
		\hline
		\multirow{5}{*}{Lowpass filter}
		& Gaussian, [3,3], std dev  0.5 & 0.017 \\
		\cline{2-3}
		& Gaussian, [3,3], std dev  1 & 0.017 \\
		\cline{2-3}
		& Gaussian,  [3,3], std dev  1.5 & 0.017 \\
		\cline{2-3}
		& Gaussian,  [3,3], std dev 2 & 0.017 \\
        \cline{2-3}
		& Averaging,  [3,3] & 0.017 \\
		\hline
		Highpass filter & unsharp, Laplacian, parameter 0.2 & 0.017 \\
		\hline
		\multirow{2}{*}{Denoising filter} & Wiener, size [3,3] & 0.05 \\
		\cline{2-3}
		& Wiener, size [5,5]  & 0.05 \\
		\hline
		Noise adding & Gaussian, 0 mean, variance 0.001 & 0.1 \\
		\hline
		\multirow{2}{*}{Tonal adjustment} & histogram stretching (saturation 2\%, shape parameter) & 0.033 \\
		\cline{2-3}
		& histogram stretching (saturation 6\%, shape parameter 0.8) & 0.033 \\
		\hline
		Histogram equalization &  --  & 0.033 \\
		\hline
		{JPEG} compression & Quality Factors (QFs) in  \{55:5:100\} & 0.1 \\
		 \hline
	\end{tabular}
	\label{table:pp_details}
\end{table}

\subsection{Parameters setting for networks training and fusion}
\label{sec:exp_settings}

The two networks {\ournet} and {\oursiam} were trained independently by using $810,000$ images from the SYN-Tr dataset; the remaining $90,000$ images were reserved for validation. We trained both networks for approximately $60$ epochs ($375,000$ iterations with batch size 128) using Adam optimizer. The learning rate was set to $10^{-4}$, and halved every $10$ epochs from epoch $40$ to improve convergence.
%
%
We used the Tensorflow framework for network training and testing (specifically, for the experiments we used Keras 2.0 on top of Tensorflow 1.8.0 and Cuda 9.0).

For score-level fusion, the weighs $w_T$ and $w_S = 1- w_T$ were set as in equation \eqref{fusion_w}. We considered several values of the constant $c$ and selected the one achieving the best fusion accuracies over the synthetic testing dataset SYN-Ts, corresponding to $c= 0.65$.

\subsection{Testing Scenarios}
\label{sec:test_scenarios}



The testing scenarios  considered for our experiments correspond to:
i) the ideal case of known binary mask with undistinguished $S$ and $T$, and known transformation, ii) the case of known binary mask only, and iii) the realistic case where everything is unknown and the mask corresponds to the output of a state-of-the-art CM detection and localization algorithm.
The first two scenarios were considered to test the disambiguation capability of the proposed approach, and the impact of possible inaccuracies introduced by the estimation of the geometric transformation. Then, in the third scenario, we assessed the performance of an end-to-end system for copy-move detection and localization that uses \textsf{DisTool} to identify the source and target regions of the copy-move.

\subsubsection{Known mask and transformation}
\label{sec.testScenario1}

In order to assess the disambiguation capability of \textsf{DisTool}, we consider the case in which both the binary localization mask (ground-truth localization mask) and the transformation $\mathcal{H}_\theta$ are given.
For these tests, we used the SYN-Ts and USCISI datasets, which provide the ground truth for the transformation matrix. From the forged images, the input patches of  the {\ournet} and {\oursiam} branches are determined as detailed in Section \ref{sec.FoA_4T} and \ref{sec.FoA_Siam}.
The two separate regions of the ground-truth  mask $P_1$ and $P_2$, to be given as input to \textsf{DisTool}, are isolated from the tampering map.

\subsubsection{Known mask only}
\label{sec.testScenario2}

In this second testing scenario, we considered the case where only the binary localization mask is known. We then used the method described in
the appendix to estimate the transformation from the binary masks of the two regions.
We tested the performance of \textsf{DisTool} on all the four datasets, namely SYN-Ts, USCISI, CASIA, and Grip.
Furthermore, we assessed the robustness of the system to post-processing  on SYN-Ts-H-PP and Grip datasets, for which processed versions of the forged images are provided.

\subsubsection{End-to-end performance}
\label{sec.testScenarioEtE}

In this scenario, we evaluated the performance of an end-to-end system for simultaneously copy-move localization
and source-target disambiguation by means of \textsf{DisTool}. For copy-move localization,
we considered  the patch-based algorithm in  \cite{Cozzolino2015},  hereafter referred to as DF-CMFD (Dense Field Copy-Move Forgery Detection), which works reasonably well under general conditions (e.g., also when the copy-moved area has a small size, or in presence of local post-processing).
In this case, a pre-processing step has to be applied to determine the two regions, $P_1$ and $P_2$, from the binary output mask provided by the localization algorithm. If more than two regions are identified, then the (1-1) condition is not met and the image is discarded.
Specifically, we first process the mask by applying a morphological opening, with a square structuring element of size $2\times 2$.
After that, we perform Connected Component (CC) analysis to label connected regions, sort them by size, and discard the images for which the ratio between the size of the third-ranked  and second-ranked regions is not small enough (the threshold is empirically set to 0.2).
The number of images retained after this stage is denoted as $OptIn$
\footnote{Note that, the above pre-processing is heuristic, however it is not of great interest in this paper since it does not have a strong impact on the applicability of the system.}.
%
%
The performance of the system are evaluated on the $OptIn$ set only.
For the opted out images, in fact, the two separated regions defining the CM operation cannot be identified, and the disambiguation system cannot be run.
Reasonably, this should be regarded to as a failure of the CM localization algorithm (more rarely, as a failure of the non ideal pre-processing step).

For this testing scenario, the results are compared with those achieved by BusterNet \cite{Wu2018}, which simultaneously aims at copy-move localization and source-target disambiguation.
For a fair comparison,  the disambiguation performance achieved by BusterNet are assessed by considering the subset of images for which two separate regions can be identified by the CM localization branch of the algorithm ($OptIn_B$). This subset corresponds to images for which the CM is correctly detected by the CM localization algorithm, and for which the (1-1) condition is matched, thus satisfying the working conditions we are assuming in this paper (as detailed in Section \ref{sec.formulation}). It is worth observing that, in some cases, BusterNet may return the same label for the two regions, that is, the regions are simultaneously labeled as source and target. In this case, the disambiguation part of the algorithm fails and the accuracy of the disambiguation is equivalent to a random choice (error probability equal to 0.5) \footnote{In \cite{Wu2018}, these kind of images are opted out and hence do not concur to determine the overall performance of the system. We believe that our approach to define the $OptIn_B$ set characterises better the performance of the algorithm, since in the presence of two distinct regions identified as part of a copy move, the disambiguation algorithm should always try to identify one of the two as the source region and the other as the target.}.
Notice that a comparison with BusterNet is not possible for the first two testing scenarios, since the method in \cite{Wu2018} is an end-to-end one providing at the same time the result of localization and disambiguation, without the possibility of taking a localization mask as input for the disambiguation part only.


\section{Experimental Results}
\label{sec.results}

As we said, we run our tests for the case of  single source and target copy-moves.
In the datasets considered for our experiments,  the number of images satisfying such condition are: 9984 out of 10000 for USCISI-CMFD, 1276 out of 1313 for CASIA-CMFD and the entire Grip-CMFD.

As evaluation metric, we considered the accuracy of the disambiguation task, computed as the ratio of correctly disambiguated copy-moves over the total number of opted-in images.
In the first and second testing scenario, the $OptIn$ set corresponds to the  set of all the images with single source and target.

\subsection{Known Mask and Transformation}

Table \ref{tab:res_1} reports the accuracy of {\ournet}, {\oursiam}, and after the final fusion step,  on SYN-Ts and USCISI.


These results confirm that  {\ournet} works very well in all the cases, but when the transformation is a rigid translation (SYN-Ts-Rigid) because the four patches are very similar. 
{\oursiam} instead works well in the presence of rigid translation, as expected, while it exhibits slightly lower performance in the presence of rotation and resizing.
The performance of {\oursiam} on USCISI are very poor, probably because most of the transformations in USCISI
include very strong rotation and resizing, and the boundaries are blended using a particular editing operation (Poison editing) \cite{Perez2003}, which has not been considered in our training sets.

Nevertheless, thanks to the final fusion step, the overall system achieves very good performance in all the cases and the loss of performance with respect to {\ournet} and {\oursiam} in their best performing scenarios is very limited. In particular, the results achieved by  \textsf{DisTool} on USCISI show that the proposed architecture works  well also under database mismatch conditions thus proving the good  generalization capability of our system.

\begin{table}[!h]
	\centering
	\caption{Accuracy ($\%$) of {\ournet}, {\oursiam} and  \textsf{DisTool} on SYN-Ts and USCISI. The ground-truth is available for both the mask and the transformation matrix.}
\setlength{\tabcolsep}{5pt}
	\begin{tabular}{| C{2.2cm} || C{0.7cm} | C{1.6cm} | C{1.7cm} | C{0.8cm}  | }
		\hline
		Dataset & $OptIn$ & {\ournet} & {\oursiam} & \textsf{DisTool} \\
		\hline
		\hline
		SYN-Ts-Rigid & $1000$ & $47.06$ & $97.00$ & $97.00$ \\
		SYN-Ts-Rot & $1000$ & $99.40$ & $93.10$ & $99.50$ \\
		SYN-Ts-Res & $1000$ & $99.30$ & $95.60$ & $99.90$ \\
		\hline
		USCISI & $9984$ & $94.48$ & $30.51$ & $91.40$ \\
		\hline
	\end{tabular}
\label{tab:res_1}
\end{table}

\subsection{Known Mask only}

\begin{figure}
	\centering
\includegraphics[width=0.99\columnwidth]{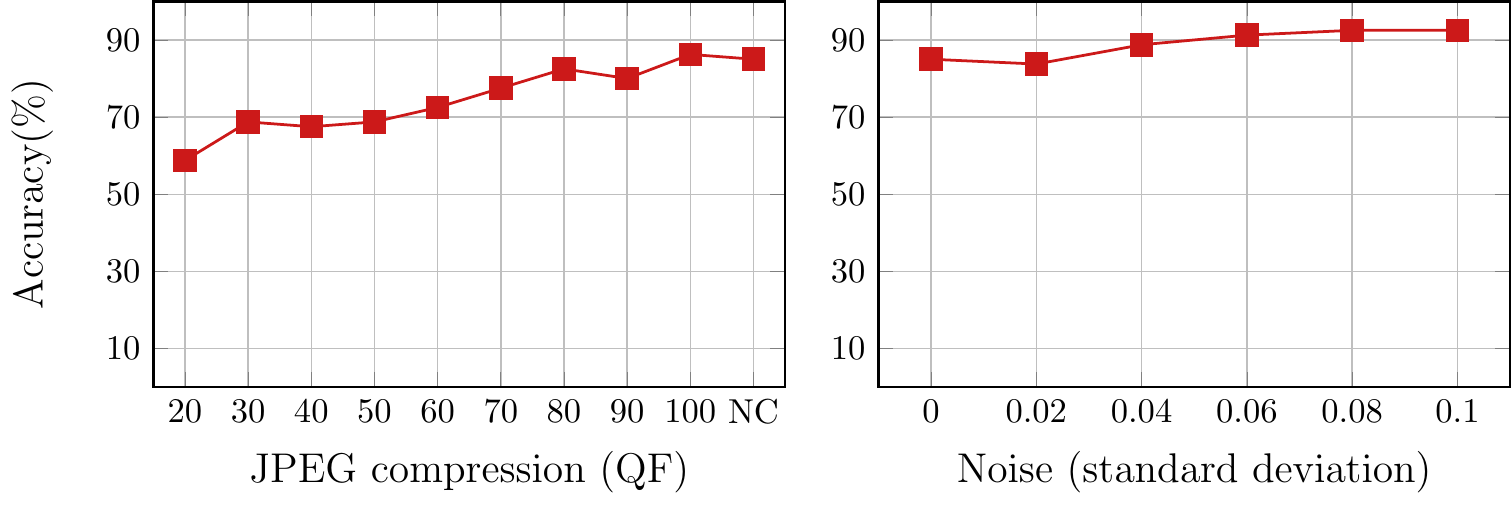}
	\caption{Accuracy of  \textsf{DisTool} (both {\ournet}, {\oursiam}, and fusion) for Grip under  JPEG compression (left) and noise addition (right). $80$ images  are considered for each case.}
\label{fig:2}
\end{figure}

The accuracies of our system in this scenario are reported in Table \ref{tab:res_2}.
%
By looking at the performance on SYN-Ts and USCISI, we can draw conclusions similar to those we drew for the known transformation case (Table \ref{tab:res_1}), thus indicating that our method for estimating the transformation works well.
When the more realistic datasets CASIA and Grip, are considered, the performance decrease a bit.
This is not surprising, given that the copy-move forgeries contained in these datasets are produced manually in different ways, and under various processing operations. For instance, forged images in CASIA are produced by Photoshop,
and advanced tools for tonal adjustments have been used. The forgeries contained in  Grip consist of visually realistic snippets designed carefully by photographic experts.
Therefore, the results achieved in these cases are also satisfactory. 
The poor performance of {\ournet} on Grip are due to the fact that the copy-move forgeries are all rigid translations (as in SYN-Ts-Rigid).
Again, the fusion step allows to improve the results of {\ournet}
in the most critical cases of close-to-rigid translations, without impairing too much the performance in the other cases.

With regard to the robustness analysis, the performance of our system in the presence of post-processing as described in Table \ref{table:pp_details}, assessed on the SYN-Ts-H-PP dataset, are reported in Table \ref{tab:res_3}.
%
%
\BTT{We also checked the robustness of our system  when global resizing is applied either before or after tampering (in which case forensic methods based on resizing detection, or more in general interpolation detection, would fail).
The results we got when the resizing is applied as a post-processing are reported in Table \ref{tab:res_PP_res}, for resizing factors equal to 0.8 and 1.2. In this case, the images in the SYN-Ts-Rigid, SYN-Ts-Rot and SYN-Ts-Res  sets are resized with the given factor then tested with \textsf{DisTool}. As it can be seen from the table, the performance remain good and the loss of accuracy is at most 1\% in the case of downsampling, and a little bit larger (less than 2\% for the resizing and rotation case, about 5\% for the rigid case)  in the case of upsampling, thus confirming that the performance of the system are not affected much by global resizing.
To verify the robustness to a global resizing performed as a pre-processing,
we purposely resized 300 pristine images in the test set and forged them as detailed in Section \ref{sec:exp_syn_db}, for both the cases of resizing, rotation and rigid CM transformations (100 for each case). The resizing factor was randomly set to 0.8. or 1.2. The results we got reveal that the average loss in the disambiguation accuracy in this case is less than 2\%.}


The robustness performance of our system on Grip dataset, under JPEG compression with different quality factors (QFs) and addition of local noise of various strength, are shown in Fig. \ref{fig:2}.
We observe that JPEG compression, being a global post-processing,  has a minor impact on the performance of the system, unless the quality of the image is significantly impaired ($QF < 60$).
Addition of local noise, instead, adds visual traces in the target and thus source and target can be more easily disambiguated when noise increases.

\begin{table}[!h]
	\centering
	\caption{Accuracy ($\%$) of {\ournet}, {\oursiam} and  \textsf{DisTool} on all the four datasets. Only the binary mask is given. }
	\begin{tabular}{| C{1.7cm} || C{0.7cm} | C{1.6cm} | C{1.7cm} | C{0.8cm} | }
		\hline
		Dataset & $OptIn$ & {\ournet} & {\oursiam} & \textsf{DisTool} \\
		\hline
		\hline
		SYN-Ts-Rigid & $1000$ & $46.00$ & $97.00$ & $97.00$ \\
		SYN-Ts-Rot & $1000$ & $98.30$ & $91.10$ & $97.90$ \\
		SYN-Ts-Res & $1000$ & $99.30$ & $95.70$ & $97.60$ \\
		\hline
		USCISI & $9984$ & $94.39$ & $45.73$ & $91.56$ \\
		CASIA & $1276$ & $69.04$ & $67.16$ & $75.86$ \\
		Grip & $80$ & $53.75$ & $86.26$ & $86.26$ \\
		\hline
	\end{tabular}
\label{tab:res_2}
\end{table}

\begin{table}[!h]
	\centering
	\caption{Accuracy ($\%$) of {\ournet}, {\oursiam} and  \textsf{DisTool} on SYN-Ts-H-PP. Only the binary mask is given.}
\setlength{\tabcolsep}{5pt}
	\begin{tabular}{| C{2.1cm} || C{0.7cm} | C{1.6cm} | C{1.7cm} | C{0.8cm}  | }
		\hline
		Dataset & $OptIn$ & {\ournet} & {\oursiam} & \textsf{DisTool} \\
		\hline
		\hline
		SYN-Ts-Rigid-PP & $1000$ & $50.40$ & $94.80$ & $94.80$ \\
		SYN-Ts-Rot-PP & $1000$ & $96.30$ & $89.60$ & $95.90$ \\
		SYN-Ts-Res-PP & $1000$ & $96.00$ & $94.70$ & $95.40$ \\
		\hline
	\end{tabular}
\label{tab:res_3}
\end{table}

\begin{table}[!h]
	\centering
	\caption{\CORR{Accuracy ($\%$) of \textsf{DisTool} on the SYN-Ts dataset, when global resizing is applied as post-processing. Only the binary mask is given.}}
\setlength{\tabcolsep}{5pt}
	\begin{tabular}{| C{1.2cm} || C{0.7cm} | C{1.7cm} | C{1.7cm} | C{1.7cm}  | }
		\hline
		Res factor & $OptIn$ & SYN-Ts-Rigid & SYN-Ts-Rot & SYN-Ts-Res \\
		\hline
		\hline
		0.8 & $1000$ & 95.90 & 97.20 & 97.30  \\
		1.2 & $1000$ & 92.50 & 96.00 & 95.10 \\
		\hline
	\end{tabular}
\label{tab:res_PP_res}
\end{table}


\subsection{End-to-end performance}
\label{res.EtoE}

In this section, we report the performance of \textsf{DisTool}  when the network is used within an end-to-end copy-move detection system with detection, localization and source-target disambiguation capabilities.
With regard to the CM detection and localization algorithm, we considered the  DF-CMFD method in \cite{Cozzolino2015} in all the cases, with the exception of the USCISI dataset, where the method in \cite{Cozzolino2015} works poorly, and  we used
the CNN-based method proposed in \cite{Wu2018} (BusterNet-CMFD).
\begin{table}[!h]
	\centering
	\caption{\CORR{Accuracy of \textsf{4-Twins} (vs  \textsf{MSE}-based disambiguator) and {\oursiam}, on all the 4 datasets for the end-to-end case (for \textsf{4-Twins} and  {\oursiam}, the CM detectors in \cite{Cozzolino2015} and
\cite{Wu2018} are considered).}}
	\begin{tabular}{| C{1.7cm} |  C{0.7cm} || C{1cm} |   C{1cm}  ||   C{1cm}  |}
		\hline
		{Dataset} & {$OptIn$} & {4-Twins accuracy} & {MSE accuracy} & {Siamese accuracy}\\
		\hline
		\hline
		SYN-Ts-Rigid  &  992  & 54.53 & 47.32  & 95.24  \\
		SYN-Ts-Rot  & 967  & 98.74 & 69.42 & 91.63 \\
		SYN-Ts-Res  & 958  & 99.21 & 86.34 & 94.45 \\
		\hline
		CASIA &  482 & 77.11 & 57.51 & 63.30  \\
		Grip &  76 & 52.32 & 59.84  & 77.63    \\
		USCISI   &  5531 & 79.80 & 40.55 & 50.48   \\
		\hline
	\end{tabular}
\label{tab:33a}
\end{table}
\begin{table*}[!h]
	\centering
	\caption{Accuracy ($\%$) of  \textsf{DisTool} for the end-to-end system on all the 4 datasets. The CM detectors DF-CMFD \cite{Cozzolino2015} and
BusterNet-CMFD \cite{Wu2018} are considered.}
	\begin{tabular}{| C{1.7cm} | C{1cm} || C{2.1cm} | C{1cm} | C{1.2cm} ||  C{1.5cm} ||  C{1cm} | C{1.2cm} || C{1.5cm} || C{1.4cm} |  }
		\hline
		\multirow{2}{*}{Dataset} & \multirow{2}{*}{\# Imags} & \multicolumn{4}{c||}{End-to-end (\cite{Cozzolino2015} / \cite{Wu2018} + \textsf{DisTool})} & \multicolumn{3}{c||}{BusterNet} & \multirow{2}{*}{\CORR{$OptIn_{Joint}$}} \\
		\cline{3-9}
		& & CM detector & $OptIn$ & $OptIn$ acc  & \CORR{$OptIn_{Joint}$ acc} & $OptIn_B$ & $OptIn_B$ acc & \CORR{$OptIn_{Joint}$ acc}  &\\
		\hline
		\hline
		SYN-Ts-Rigid & $1000$ &  DF-CMFD & $992$ & $94.86$ & \CORR{93.67} &  $143$ & $80.99$ & \CORR{80.32} & \CORR{142}\\
		SYN-Ts-Rot & $1000$ & DF-CMFD & $967$ & $98.66$  & \CORR{100} &  $33$ & $84.85$ & \CORR{83.87} & \CORR{31} \\
		SYN-Ts-Res & $1000$ & DF-CMFD & $956$ & $96.75$  & \CORR{95.83} &$146$ & $86.99$ & \CORR{86.81} & \CORR{144} \\
		\hline
		CASIA & $1276$ & DF-CMFD & $482$ & $74.04$  &  \CORR{74.72} & $688$ & $52.18$ & \CORR{15.84} & \CORR{241} \\
		Grip & $80$ & DF-CMFD & $76$ & $74.67$ &  \CORR{73.71} & $21$ & $42.86$  & \CORR{42.17} & \CORR{19}\\
		USCISI & $9984$ & BusterNet-CMFD & $5531$ & $78.61$  &  \CORR{77.65} &  $5051$ & $85.57$ & \CORR{86.44} & \CORR{4179} \\
		\hline
	\end{tabular}
\label{tab:3}
\end{table*}
%
\begin{table*}[!h]
	\centering
	\caption{\CORR{Performance of tampering localization of the end-to-end DisTool (for the same CM detectors in \cite{Cozzolino2015} and \cite{Wu2018}) and BusterNet.}}
	\begin{tabular}{| C{1.7cm} || C{1.3cm} ||  C{0.9cm} |   C{0.8cm}   | C{1.2cm} | C{0.9cm} |   C{0.8cm} |  C{1.2cm}   ||  C{0.9cm} |   C{0.8cm} |  C{1.2cm}   |}
		\hline
\multirow{2}{*}{Dataset}  &  \multirow{2}{*}{$OptIn_{Joint}$} & \multicolumn{3}{c|}{\textsf{DisTool} end-to-end } & \multicolumn{3}{c||}{baseline CM detector (\cite{Cozzolino2015} / \cite{Wu2018})}   & \multicolumn{3}{c|}{BusterNet} \\
\cline{3-11}
& &  {Precision}& {Recall}  & {F1-score} & {Precision}& {Recall} & {F1-score} & {Precision}& {Recall} & {F1-score} \\
		\hline
		\hline
		SYN-Ts-Rigid &   142  & 0.899 & 0.915 & 0.906 & 0.961 & 0.487 & 0.646 & 0.375 & 0.659 & 0.439 \\
		SYN-Ts-Rot &  31  &  0.871 & 0.953   & 0.899 & 0.910 & 0.493 & 0.632 & 0.485 & 0.739  & 0.552   \\
		SYN-Ts-Res &  144  &  0.906 & 0.938 & 0.920 & 0.946 & 0.507 & 0.658 & 0.500 & 0.734  & 0.563  \\
		\hline
		CASIA &  241  & 0.659 & 0.644  & 0.643 & 0.892 & 0.419 & 0.565 & 0.126 & 0.357 & 0.163  \\
		Grip &   19  & 0.650 & 0.715  & 0.679 & 0.895 & 0.457 & 0.600 & 0.009 & 0.053  & 0.016 \\
		USCISI &  4179  & 0.405 & 0.729   & 0.486 &  0.518 & 0.293 & 0.339 &  0.439 & 0.766 & 0.507  \\
		\hline
	\end{tabular}
\label{tab:33e}
\end{table*}
In this case, the $OptIn$ images are the images for which the two duplicated regions can be correctly identified after the application the CM localization algorithm and the pre-processing.

\BTT{The results of our tests are reported in Table \ref{tab:33a} and \ref{tab:3}.
The separate accuracies of {\ournet} and {\oursiam} in this more general scenario are reported in Table \ref{tab:33a}, while the final accuracies of \textsf{DisTool} are reported in Table \ref{tab:3},
together with those of BusterNet.
In particular, in Table \ref{tab:33a}, the performance of {\ournet} are compared to those achieved by the MSE-based disambiguator, where the source-target decision is done by looking at the pair with the lowest MSE among the two input pairs (see Section \ref{sec.4Twins}).
The fact that the number of $OptIn$ images is lower than the total number, much lower in some cases, is mainly due to the failures of the CM detection algorithm.}

\CORR{
From Table \ref{tab:33a}, we see that the accuracy obtained with an MSE-based disambiguator is significantly lower than the accuracy of {\ournet} for all the datasets and, in general, MSE-based decision works very poorly in all cases except for the synthetic dataset and non-rigid transformations, i.e. for SYN-Ts-Rot and in particular SYN-Ts-Res.
Expectedly, all the non-idealities of this scenario, namely, the fact that the localization mask is not ideal and the transformation is an estimated one, affect significantly the performance of a simple MSE-based decision.}

In  Table  \ref{tab:3}, the performance of \textsf{DisTool} are compared to those achieved by BusterNet.
Even if the number of $OptIn$ and $OptIn_B$ images is not the same, mainly due to the difference in the localization method adopted in the two cases, $OptIn$ and $OptIn_B$ have a similar meaning (see Section \ref{sec.testScenarioEtE}),
then,  the disambiguation performance achieved by our system on the $OptIn$ set can be compared to those achieved by BusterNet on the set of $OptIn_B$ images. \CORR{However, for a more fair comparison, we also report the performance achieved by both methods on the subset of opted-in images common to \textsf{DisTool} and BusterNet, that is $OptIn \cap OptIn_B$, indicated as $OptIn_{Joint}$ in the table. The number of images in the $OptIn_{Joint}$ set for the various datasets is reported in the last column.}

Noticeably, the number of duplicated regions correctly localized  by DF-CMFD ($OptIn$) is always higher than the corresponding number by BusterNet ($OptIn_B$),
the only exception being CASIA, where $OptIn_B$ is 688, while $OptIn$ is 482. However, the performance of BusterNet in this case are very poor, and the average accuracy of the disambiguation is about 50\% (hence similar to a random guess).
%
%
We also observed that several times the same label (source or target) is assigned to the two duplicated regions by  BusterNet, meaning that the method is not able to disambiguate between them.
\CORR{To be more specific, only for 126 images over 688, BusterNet assigns different labels to the source and target regions (with a disambiguation accuracy 76.19\%), while in all the remaining cases the same label is assigned to both source or target. Something similar happens with Grip, where different labels are assigned to the regions for only 3 images (out of 21), with a wrong decision in all the cases. }
%
%
%
%
By inspecting the table, we see that BusterNet gives better results compared to our method only on USCISI, which is the same dataset used for training, hence corresponding to a favorable case for that method. 
With the exception of the USCISI dataset,  \textsf{DisTool} always outperforms BusterNet, achieving a better accuracy on all the datasets. Noticeably, \textsf{DisTool} works pretty well in the most difficult cases with public realistic datasets (CASIA and Grip).
To assess the robustnesss of  \textsf{DisTool} against post-processing in the realistic scenario of CM forgery we post-processed the images in the CASIA dataset and tested \textsf{DisTool} on them. As for the synthetic case, a resizing is applied to the tampered images with scale factor 0.8 and 1.2. The $OptIn$ accuracies that we obtained in the two cases are respectively  71.61 and  75.31 (the number of opted-in images  in the two cases is 310 and 563 respectively).

\CORR{For completeness, in Table \ref{tab:33e}  we report the results of CM tampering localization achieved by using \textsf{DisTool} on top of the CM detectors in \cite{Cozzolino2015} and \cite{Wu2018}, compared to the end-to-end BusterNet system, on the  $OptIn_{Joint}$ set.  Localization results are provided by letting the final tampering mask correspond to the target region identified after the disambiguation step. In this way only the pixels that have been actually modified by the copy-move operation are assumed to be tampered. For completeness, the performance of the CM detector, called baseline, are also reported in the table. In this case, the final tampering mask contains both the source and target regions.
We see that the localization performance of the proposed scheme are by far better than those achieved by BusterNet,
due to the poor localization capability of such a network, thus confirming the advantages of having an independent tool for disambiguation, that can exploit existing well performing algorithms for CM detection and localization.}
With regard to the CM detector, expectedly, the precision values obtained after disambiguation are worse than those achieved by the baseline detector, since, a wrong identification of the source and target regions, sometimes prevents a correct localization of the tampered pixels. However, this corresponds to a significant increase of the recall, eventually resulting in a much better F1-score.

We further emphasize that, in our experimental analysis, we only considered two common and well-known methods  for CM detection and localization, namely those in  \cite{Cozzolino2015} and in \cite{Wu2018}. Other methods could be considered as well (for instance \cite{Li2018}).
The fact that our systems can work on top of any CM localization algorithm is in fact a remarkable strength of the approach. Moreover, since different methods (e.g. patch-match based or keypoints-based) have often different peculiarities and
work better in different conditions,
the best CM localization method could be chosen based on the kind of images under analysis.

\section{Concluding  Remarks}
\label{sec.conc}

We have proposed a method for  source-target disambiguation in copy-move forgeries.
This problem has not gained much attention in the past, yet solving the disambiguation problem is of primary importance to correctly localize the tampered region in a copy-move forgery. Common existing algorithms, in fact, identify both the original (source) and copied (target) region, yet  only the target region corresponds to a tampered area.
To address this problem, we leveraged on the capability of deep neural network architectures to learn suitable features for exposing the target region, by looking at the presence of interpolation artefacts and boundary inconsistencies. Specifically, we proposed an architecture
with two multi-branch CNNs that extract different features and perform the disambiguation independently; then, decision fusion is applied at the score level.
Our experiments show that our disambiguation method, called \textsf{DisTool}, performs well even in the realistic testing scenario, where the copy-move binary localization mask is provided by a CM detection algorithm and the CM transformation is estimated from such mask.
Based on our tests, the proposed architecture trained on a synthetic dataset achieves good results also on copy-move images from realistic public datasets, then the generalization capability of the method is also good.

%
As a future work, we could investigate other strategies to perform fusion of the network outputs. In particular, methods based on machine learning could be adopted, e.g. an SVM or a random forest classifier. 
Another interesting possibility would be to resort to fuzzy logic fusion \cite{FuzzyBook}. A fuzzy fusion module could also be integrated in the multi-branch CNNs architecture, that could then be trained as a whole. In this way, the weights of the fuzzy logic module could also be optimized through backpropagation \cite{cho1995multiple}.
%
\CORR{The analysis of the multi-target copy moves scenario
could also be considered as future research. In this case, a  pre-processing could be carried out
to trace back the problem to the solution of several ($1$-$1$) problems,
that can then be solved using the \textsf{DisTool} architecture presented in this paper.}

\CORR{Finally, by adopting the forger's perspective, we could evaluate the robustness of DisTool and see
if it can be fooled by an informed or partially informed attacked, e.g. via adversarial examples or by means of some ad-hoc processing applied to the source or target region (e.g. applying blurring to the source), and assess the amount of distortion that need to be introduced to make the system fail.}
\BTT{Related to this point, it could be also interesting to perform a deep analysis of the layer activation maps and the input saliency maps to investigate what the networks are actually learning and what is the main focus of their analysis.}

\section*{Acknowledgements}

This work has been partially supported by a research sponsored by DARPA and Air Force Research Laboratory (AFRL) under agreement number FA8750-16-2-0173. The U.S. Government is authorised to reproduce and distribute reprints for Governmental purposes notwithstanding any copyright notation thereon. The views and conclusions contained herein are those of the authors and should not be interpreted as necessarily representing the official policies or endorsements, either expressed or implied, of DARPA and Air Force Research Laboratory (AFRL) or the U.S. Government.

The authors would also like to thanks Giulia Boato from the University of Trento for advice and financial support.

\appendix


\subsection*{Geometric Transformation Estimation.}

\CORR{For sake of experimental reproducibility, we provide a detailed  description of the geometric estimation procedure adopted in our system.
\begin{figure}
	\center
	\includegraphics[width=0.85\linewidth]{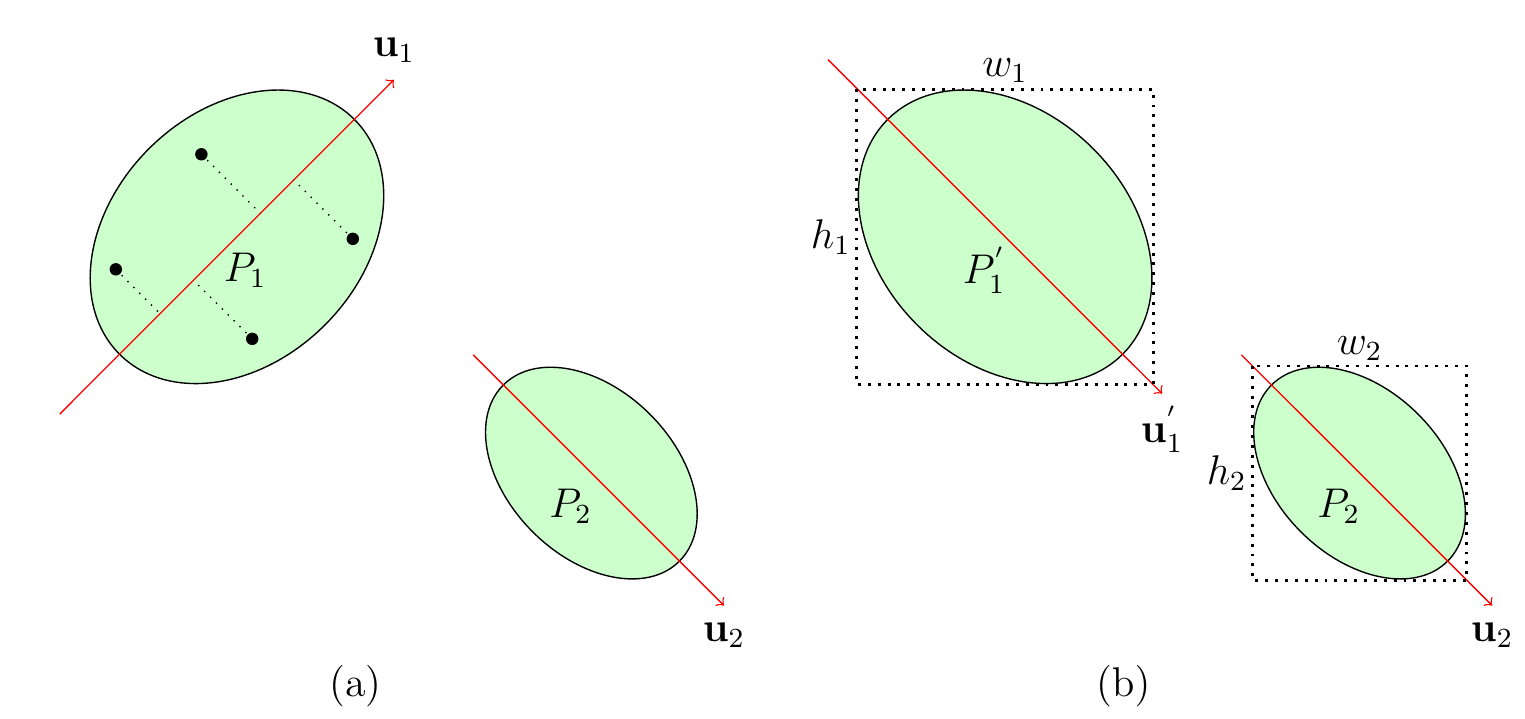}
	\caption{Illustrations of rotation angle and scaling factor estimation.}
	\label{fig:trans_est}
\end{figure}
Given the binary localization mask, we want to estimate the homography matrix
that maps the pixels in $P_1$ into those of $P_2$, to obtain a remapped region $\widetilde{P}_2$ that is as similar as possible to $P_2$ (and viceversa for the backward transformation). In its more general form, ${\cal H}_{\theta}$  is an affine geometric transformation.\footnote{With a slight abuse of notation, regions $P_1$ and $P_2$ are regarded as the sets with the coordinates of the pixels in the $x-y$ image plane, and not as the values of the pixels belonging to the regions.}
%
In this paper, we consider only {\em similarity}
transformations, namely translations,  resizing or scaling and rotations
and, more in general, a composition of them. In this case, the transformation
can always be expressed as the subsequent application of a rotation with angle $\alpha$, a scaling with  factors $f_x, f_y$, and two translations $t_x, t_y$,
represented in homogenous coordinates by the following matrix
\footnote{While, in general, any composition of translations, rotations and scalings can be perfectly represented by a single rotation followed by a single scaling and a single translation, the interpolation artefacts associated to real geometric transformations may introduce some dependency on the order and the exact way the different transformations are applied. In this paper we neglect such phenomenum.}
\begin{align}
\label{H123}
H_\theta 
= &
	\begin{bmatrix}
		1 & 0 & t_x \\
		0 & 1 & t_y\\
		0 & 0 & 1
	\end{bmatrix}
	\begin{bmatrix}
		f_x & 0 & 0 \\
		0 & f_y & 0\\
		0 & 0 & 1
		\end{bmatrix}
	\begin{bmatrix}
		\cos(\alpha) & - \sin(\alpha) & 0 \\
		\sin(\alpha) & \cos(\alpha) & 0\\
		0 & 0 & 1
		\end{bmatrix}.
\end{align}
%
The estimation of the parameters is carried out according to the following three steps: i) estimation of the rotation angle $\alpha$, ii) estimation of the resizing factors $f_x, f_y$, and iii) estimation of the translations $t_x$ and $t_y$.}

\CORR{To estimate $\alpha$, we find the two central principal inertia axes of ${P}_1$ and ${P}_2$ and let $\alpha$ be equal to the difference between them.
In particular, we adopt a Principle Component Analysis (PCA) \cite{Hotelling1933}, to determine the direction along with the second-order central moments of projected points is maximized.
%
%
The directions found in this way, represented by the column vectors ${\bf u}_1 = [u_{1,1}, u_{1,2}]^T$, and ${\bf u}_2  = [u_{2,1}, u_{2,2}]^T$, are illustrated in Fig. \ref{fig:trans_est}(a).
More specifically, let us denote with $\bf P$ the $2 \times N$ matrix whose $i$-th column ${\bf p}_i$ represents the vector of 2-D coordinates of point $i$ within ${P}_1$, $1 \leq i \leq N$, where $N$ denotes the number of points in ${P}_1$,
and with $\bar{p} = \frac{1}{N} \sum_{i=1}^{N} {\bf p}_i$ the centroid of ${P}_1$. If we assume ${\bf u}_1^T {\bf u}_1 = 1$, the second-order central moment of the projected points in $P_1$ is given by:
\begin{equation}
	\frac{1}{N} \sum_{i=1}^N ( {{\bf u}^T_1\bf p}_i - {\bf u}^T_1\bar{p} )^2 = {\bf u}_1^T {\bf S}_1 {\bf u}_1 \text{,} \nonumber
\end{equation}
where ${\bf S}_1 = \frac{1}{N} \sum_{i=1}^N \left( {\bf p}_i - \bar{p} \right) \left( {\bf p}_i - \bar{p} \right)^T$ is the inertia matrix of the points in ${P}_1$. The principle component ${\bf u}_1$ is obtained as:
\begin{equation}
	{\bf u}_1   = \arg \underset{{\bf u} \text{:} {\bf u}^T {\bf u} = 1}{\max} \; {\bf u}^T {\bf S}_1 {\bf u} \;  \text{.}  \nonumber
\end{equation}
We find ${\bf u}_2$ in a similar way.
Then the angle $\alpha$ is computed as: $\alpha = \tan^{-1} \left\{ \frac{{u}_{2,2}}{{u}_{2,1}} \right\} - \tan^{-1} \left\{ \frac{{u}_{1,2}}{u_{1,1}} \right \}$.
%
%
Once the rotation angle $\alpha$ has been estimated,
${P}_1$ is rotated by $\alpha$ thus obtaining a new region ${P}_1^{'}$. To estimate the scaling parameters $f_x$ and $f_y$, we first determine the $h_1 \times w_1$ bounding box of ${P}_1^{'}$ and the $h_2 \times w_2$ bounding box of ${P}_2$, as illustrated in Fig. \ref{fig:trans_est} (b), the scaling factors are then computed as: $f_x = \frac{w_2}{w_1}, f_y = \frac{h_2}{h_1}$. Finally, the translation terms are merely the difference between the centroids of ${P}_1^{'}$ and ${P}_2$.}

\CORR{By applying the same procedure to estimate the transformation mapping $P_2$ into $P_1$, we would simply obtain ${H}_{\theta}^{-1}$. Therefore, for simplicity, the transformation is estimated in one direction only (as detailed above), and ${H}_{\hat{\theta}}^{-1}$ is used as the transformation bringing $P_2$ onto $P_1$\footnote{This is possible since the estimated transformation
applies to the points in the binary mask, and not directly to the pixel values (that is, interpolation of grey levels is not considered to estimate ${H}_{\hat{\theta}}$).}.}

\ifCLASSOPTIONcaptionsoff
\newpage
\fi
\bibliographystyle{IEEEtran}
\bibliography{references.bib}

\end{document}